\documentclass[sn-basic,iicol, pdflatex]{sn-jnl}% Math and Physical Sciences Reference Style
\jyear{2021}%

\theoremstyle{thmstyleone}%
\theoremstyle{thmstyletwo}%
\theoremstyle{thmstylethree}%

\usepackage{graphicx}
\usepackage{amsmath,bm}
\usepackage{amssymb}
\usepackage{booktabs}
\usepackage{bm}
\usepackage{multirow}
\usepackage{arydshln}
\usepackage{footmisc}
\usepackage[T1]{fontenc}
\usepackage{caption}
\usepackage{enumitem}
\usepackage{xcolor}
\newcommand{\doublecheck}[1]{\textcolor{black}{#1}}
\newcommand{\keypoint}[1]{\vspace{0.02cm}\noindent\textbf{#1}\;}
\DeclareMathOperator{\LogSumExp}{LogSumExp}
\DeclareMathOperator{\SoftPlus}{SoftPlus}

\DeclareMathOperator{\argmax}{argmax}
\DeclareMathOperator{\onehot}{one\_hot}
\DeclareMathOperator{\Gumbel}{Gumbel}
\DeclareMathOperator{\Uniform}{Uniform}
\DeclareMathOperator{\stopgradient}{stop\_gradient}
\def\eg{\emph{e.g., }}
\def\ie{\emph{i.e., }}
\def\vs{\emph{vs. }}
\def\cf{\emph{cf. }}

\begin{document}

\title[Sketch Quality Assessment]{Annotation-Free Human Sketch Quality Assessment}


\author[1,2]{\fnm{Lan} \sur{Yang}}\email{ylan@bupt.edu.cn}

\author[2]{\fnm{Kaiyue} \sur{Pang}}\email{thatkpang@gmail.com}

\author*[1]{\fnm{Honggang} \sur{Zhang}}\email{zhhg@bupt.edu.cn}

\author[2]{\fnm{Yi-Zhe} \sur{Song}}\email{y.song@surrey.ac.uk}

\affil[1]{\orgdiv{PRIS Lab}, \orgname{School of Artificial Intelligence, Beijing University of Posts
and Telecommunications}, \orgaddress{\city{Beijing}, \postcode{100876}, \country{China}}}

\affil[2]{\orgdiv{SketchX Lab}, \orgname{Centre for Vision, Speech and Signal Processing, University of Surrey}, \orgaddress{\city{Surrey}, \postcode{GU27XH}, \country{United Kingdom}}}


\abstract{As lovely as bunnies are, your sketched version would probably not do them justice (Fig.~\ref{fig:intro}). This paper recognises this very problem and studies sketch quality assessment for the first time -- letting you find these badly drawn ones. Our key discovery lies in exploiting the magnitude ($L_2$ norm) of a sketch feature as a quantitative quality metric. We propose Geometry-Aware Classification Layer (GACL), a generic method that makes feature-magnitude-as-quality-metric possible and importantly does it without the need for specific quality annotations from humans. GACL sees feature magnitude and recognisability learning as a dual task, which can be simultaneously optimised under a neat cross-entropy classification loss with theoretic guarantee. This gives GACL a nice geometric interpretation (the better the quality, the easier the recognition), and makes it agnostic to both network architecture changes and the underlying sketch representation. Through a large scale human study of 160,000 \doublecheck{trials}, we confirm the agreement between our GACL-induced metric and human quality perception. We further demonstrate how such a quality assessment capability can for the first time enable three practical sketch applications. Interestingly, we show GACL not only works on abstract visual representations such as sketch but also extends well to natural images on the problem of image quality assessment (IQA). Last but not least, we spell out the general properties of GACL as general-purpose data re-weighting strategy and demonstrate its applications in vertical problems such as noisy label cleansing. Code will be made publicly available at github.com/yanglan0225/SketchX-Quantifying-Sketch-Quality.}




\maketitle

\section{Introduction}\label{introduction}

\begin{figure*}
    \centering
    \includegraphics[width=\textwidth]{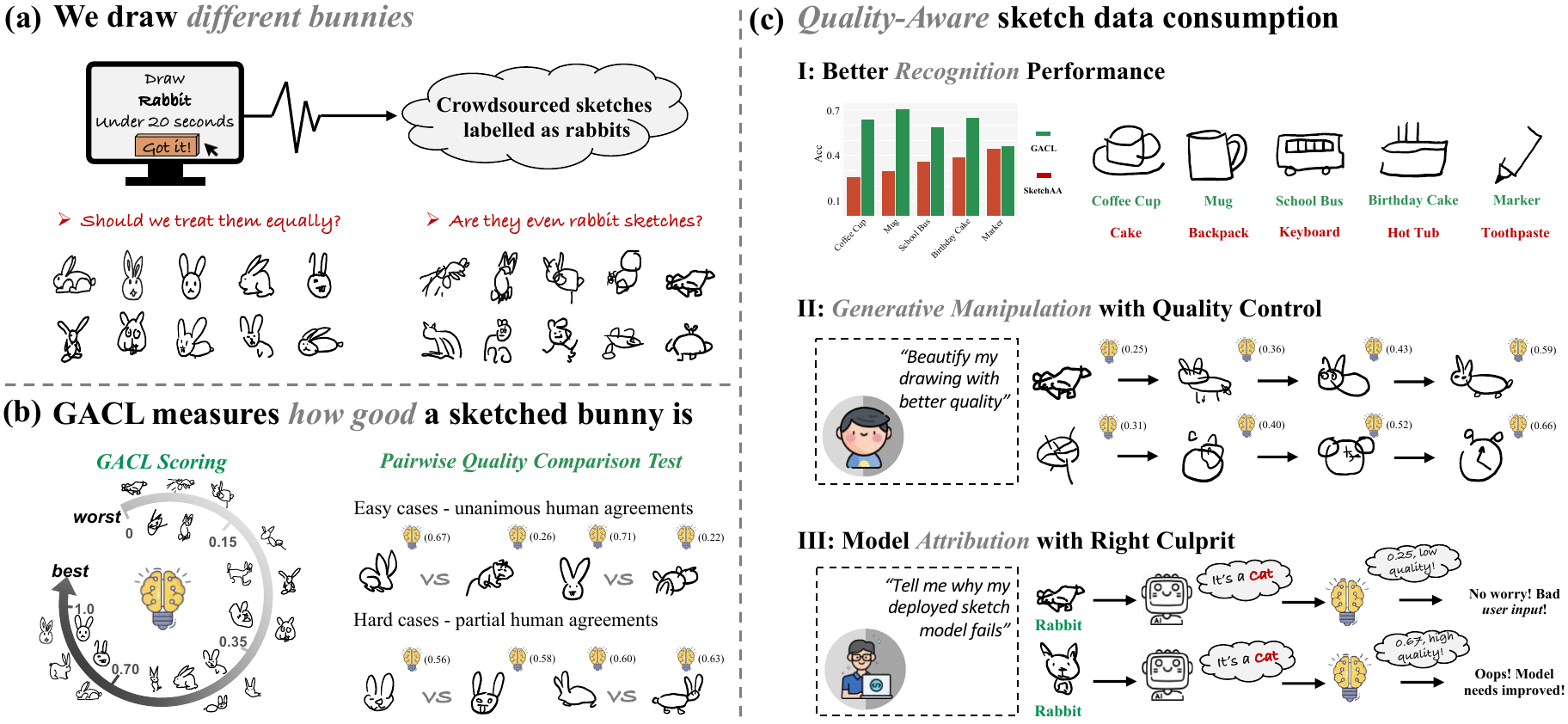}
    \vspace{0.05cm}
    \caption{\textit{Motivation and Overview.} (a) Not every free-hand sketching bunny is of equal quality. (b) We contribute a solution (GACL) for discriminating quality between bunny (and many other categories) sketches. Without the need of collecting human quality ground truth for training, we show quality discovery under GACL supports reasonable level of quality examination from humans. (c) We demonstrate three practical applications benefited from quality-aware sketch data consumption (SketchAA \citep{sketchaa} is the latest work representing state-of-the-art sketch recognition performance).}
    \label{fig:intro}
\end{figure*}

Everybody \textit{can} sketch, the debate is on \textit{how well}. With the proliferation of touchscreen devices, the urge to sketch is ever more pronounced. This is not to mention the broad range of sketch-enabled applications -- from recognition \citep{eitz2012humans,schneider2014sketch,sketchanet,lin2020sketch}, parsing \citep{sarvadevabhatla2017sketchparse,li2018universal,lumin2021sketchgnn}, recreating \citep{quickdraw,liu2019sketchgan,sketchhealer,das2021cloud2curve}, to leveraging sketch as a query modality for image search \citep{sangkloy2016sketchy,pang2019generalising,liu2020scenesketcher,bhunia2020fgsbir,sain2021stylemeup} and visual content manipulation \citep{zhu2016generative,chen2020deepfacedrawing,wang2021sketch,jo2019sc}, from facilitating product designs \citep{gryaditskaya2019opensketch} to even enabling a drawing agent that excels human at a Pictionary-like sketching game \citep{bhunia2020pixelor}.

This paper recognises this very ``how well'' problem and proposes a learnable metric that for the first time tells us just \textit{how badly} drawn my bunny (or any other sketch) is -- so from the collection of bunny sketches in Fig.~\ref{fig:intro}(a) to an ordered list of bunnies from worse to best in Fig.~\ref{fig:intro}(b). As interesting as the problem sounds in its own right, it also underpins many facets of sketch research at large. These include but are not limited to \doublecheck{
(i) facilitating better representation learning, \ie making data-driven models less prone to overfitting by learning against specific quality level; (ii) sketching assistance applications, \eg helping users to move towards a better bunny sketch and even leveraging it as a vehicle for better bunny image generation; (iii) disentangling human factor in model prediction, \ie whether the model is bad or the culprit is the sketch.}

Quantifying sketch quality is non-trivial. The first obstacle is the fatal lack of existing sketch datasets annotated with human quality ratings. This essentially renders most of the recent works on image quality assessment that predicts human opinion scores inapplicable \citep{rankiqa,talebi2018nima,zhu2020metaiqa,she2021hierarchical}. The sketch-specific trait as a vector representation of sequential coordinates also sets it apart from another line of works on trying to model human-interpretable image quality directly from low-level statistical distortions \citep{wang2004image,kim2017deep,zhang2018unreasonable,ahn2021deep} -- compared with visual artefacts such as noise, blur and compression, sketch quality is more of a subjective interpretation of holistic visual concepts.

In this paper, we provide the first stab at learning a sketch-specific quality score (metric) \textit{without} the reliance on human quality annotations. Core to our technical solution is the insight that such a score is inherent to the feature space geometry underlying a recognition task (Fig.~\ref{fig:relation}). We first single out the orderless feature geometry learned off a typical Softmax formulation (Eq.~\ref{eq:nsoftmax}) to be the main culprit for the failed quality discovery. This is because Softmax will constantly push sketch features to be close to the class centre and thus undermine any potential geometry formation. The intuition is then that a real-valued feature magnitude can already induce a quality metric, \textit{if} the underlying feature geometry can satisfy the following property: the better quality a sketch, the closer its feature to the class centre.

We incorporate such insight by making theoretic modifications upon Softmax and propose a quality-informed alternative named Geometry-Aware Classification Layer (GACL). GACL implements quality score as feature magnitude ($L_{2}$ norm) and treats its learning alongside recognition as a dual task. This then importantly gives rise to the geometry constraints said above, where the monotonic increase of sketch quality score is positively correlated with easy recognisability, \ie closer to a class centre. To encourage the integrity of learned feature magnitude, we enforce its optimisation to be convex and that a global \doublecheck{optimum} is guaranteed. We also show that under mild mathematical approximation, the quality score of a sketch instance under GACL equates to the distance to its class decision boundary, yielding nice semantics to the quality defined (that the better the quality, the farther away a sketch is from its class decision boundary). 

{The vanilla GACL recipe above, which appeared in an earlier version of this work \citep{lancvpr}, performs reasonably well in practice, however with one important caveat -- sketches of strong semantically-tied clusters (\eg head-only bunnies) are always deemed higher quality than others (\eg full-body bunnies), being oblivious to the fact that certain full-body ones could in fact be better than head-onlys (see Fig.~\ref{fig:semantic}(a)). A closer inspection reveals this is part of a more systematic oversight -- we had taken the assumption that recognisability learning and quality scoring are orthogonal tasks, and this has proven to be an unsubstantiated assumption. The very reason behind the undesired phenomenon in  Fig.~\ref{fig:semantic}(a) is because recognisability learning had come to early conclusions that frontal-bunnies are more discriminative than those depicted from the side -- this renders frontal-bunnies to be tightly clustered before a reasonable quality score is assigned. The direct result is therefore when a quality score is given, it would consider all tightly sitting frontal-bunnies in the feature space to be of high/low quality. Our solution is intuitive -- we refrain from assigning quality scores until after reasonable class boundaries are formed. This essentially forces the model to be more ``instance-aware'' as discriminability learning becomes more fine-grained \citep{arpit2017closer, baldock2021deep, hu2020surprising, kalimeris2019sgd}. More specifically, we resort to the clever use of the soft binning trick \citep{dougherty1995supervised, yang2018deep} to allow GACL being still fully differentiable for end-to-end optimisation. By binning the continuous quality scores into a pre-defined ordered set of value buckets, we give quality learning a ``slack off'' time by stopping to proceed for discriminating quality scores within the same bucket. We verify the efficacy of our proposal in Fig.~\ref{fig:semantic}(b)(c).}

{We develop four specific instantiations of GACL and conduct human studies respectively to provide some assurance on the ordering of the quality scores learned. Asking humans to annotate an exact score for a given sketch/image is however not feasible \citep{liang2014beyond, fu2015robust}, as subjectivity varies to a large degree. For that we undertake a ``pairwise comparison'' strategy that is commonly adapted in the psychology literature as an alternative approximation \citep{pairwise, kendall1940method}. More specifically, we ask human participants to judge between two sketches which is better (in a relative sense), instead of assigning an absolute quality score independently to each. We undertake extensive pairwise trials to best simulate a global approximation from the pairwise annotations \citep{liu2009learning, he2022gnnrank}. Results off 160,000 trials (40 participants each doing 4,000 trials) have humans agreeing with the learned quality ordering 78.13\% of the time across 8 carefully selected sketch categories.}
    
{Importantly, we showcase the practical benefits of modelling sketch quality under three novel sketch applications that \doublecheck{would otherwise be challenging to realise}: (i) a quality-aware sketch recognition model contributes to the new state-of-the-art recognition performance; (ii) a quality-guided sketch generative model pushes the envelope of sketch manipulation task beyond generating conceptually correct sketches; (iii) a quality-enabled sketch model attribution method that helps sketch practitioners to identify malicious user input. The benefits of GACL however are not constrained to sketch data domain. We wrap up the paper with in-depth analysis confirming that GACL does not only achieve comparable or SoTA performance on traditional image quality assessment task across six benchmarks, but is also a promising general vision data re-weighting method.}

\section{Related Work}
\label{related work}

\keypoint{Sketch research.} Apart from constantly raising the performance bar on various sketch perceptual tasks, recent computer vision works for human sketch data have been additionally focusing on two unique aspects: (i) pixel/vector dichotomy: should sketch be processed as raster pixel image \citep{sketchanet,sangkloy2017scribbler,pang2020solving,wang2021sketch} or vector graphic \citep{quickdraw,muhammad2018learning,song2018learning,lin2020sketch} compiled as a sequence of points, or the combination of both \citep{xu2018sketchmate,sketchhealer,wang2021sketchembednet,sketchaa}? Current explorations suggest that better performance is often obtained when the two modalities are encoded cooperatively as one unified representation for either generative or discriminative tasks. (ii) ``can't sketch'' reality: unlike clicking a tag or typing a search keyword, sketching is a slow and skilful process. Users can be worrying about inaccurate results because of their poor renderings and consequently not motivated enough to sketch in the first place. Existing solutions include allowing users to stop early in a sketching episode so that their goals can be achieved with earliest/easiest strokes \citep{lee2011shadowdraw, bhunia2020fgsbir, bhunia2020pixelor,bhunia2022sketching} or a real-time drawing assistant that lowers rendition barriers \citep{xie2014portraitsketch,matsui2016drawfromdrawings,shi2020emog}. We study a new sketch problem of computational quality modelling, which can potentially benefit many ongoing sketch research -- from improving discriminative performance (Sec.~\ref{sec:recog}) to introducing a beautification objective into existing generative models (Sec.~\ref{sec:generate}).

\keypoint{Image quality assessment.} Existing literature on image quality assessment (IQA) draws a distinction between approaches that require an input reference and those do not. Referenced-based algorithms \citep{kim2017deep,zhang2018unreasonable,hammou2021egb} assume the availability of pristine and distorted image pairs so that the quality gap can be measured, where the no-reference or blind IQA \citep{zhang2015feature,zhang2021uncertainty,wang2021troubleshooting} loosens the pairing constraint by instead exploiting from a carefully curated image set processed with several known distortion types (\eg noise, blurring, corruptions, and compression artefacts). One particular line of blind IQA works \citep{gao2015learning,talebi2018nima,zhu2020metaiqa, ma2017dipiq} is how to accurately predict subjective human quality ratings provided by datasets like AVA \citep{murray2012ava} and LIVE \citep{ghadiyaram2015massive}, which is not applicable here due to the lack of a similarly annotated sketch dataset. We too approach sketch quality assessment as a blind IQA problem and propose a novel solution to help us to bypass the otherwise laborious and expensive human annotation step. We also show our proposal excels at the traditional IQA benchmarks when re-purposed properly.
                    
\keypoint{Margin-based learning.} Margin is an important concept for representation learning before the deep learning wave (\eg SVM \citep{cortes1995support} is also known as soft margin classifier), and even more so when deep learning sweeps across computer vision fields today (\eg contrastive \citep{hadsell2006dimensionality} or triplet ranking loss \citep{wang2014learning}). Most relevant to ours is the idea of encapsulating margin into a Softmax-based classification head. By modifying the vanilla Softmax via the insertion of either fixed or adaptive margin, many representative Softmax variants have been proposed \citep{normface,sphereface,cosface,arcface,liu2019fair,magface} to boost feature discriminativeness with the same goal of ensuring within-class variation is smaller than between-class difference. We have shown analytically that quality score learned under our framework corresponds to the instance-specific margin to the class decision boundary, and thus give an intuitive explanation of the GACL inner workings from a feature space geometry view (Sec.~\ref{sec:explain}).

\section{Methodology}

The goal of this paper is to obtain a score-based metric $q(\cdot)$ that quantifies sketch quality. Given sketch sample $x_i$ with category label $y_i\in \{1,2,..., C\}$, our key finding is that, during the training of a sketch recognition network $f(\cdot)$ and under certain mild conditions, sketch feature magnitude ($L_2$ norm) can automatically encode the computational metric $q(\cdot)$ needed for quality discrimination, \ie $q_i\equiv q(x_i)= \parallel f(x_i)\parallel_2$. We will first introduce the necessary preliminaries before describing our proposed method for $q(\cdot)$ to be a good proxy for quality discovery.

\subsection{Preliminaries and Discussions}

In a conventional Softmax-based classification layer, the training objective for a sample $x_i$ being classified as its ground truth category $y_i$ is formulated as:
\begin{equation}
\label{eq:softmax}
L_{sm}(x_i)= -\log\frac{e^{W_{y_i}^{T}f(x_i)+B_{y_i}}}{e^{W_{y_i}^{T}f(x_i)+B_{y_i}}+\sum\limits_{j=1, j\neq i}^C e^{W_{y_j}^{T}f(x_i)+B_{y_j}}} 
\end{equation}

\noindent where $f(x_i) \in \mathbb R^d$ is the extracted deep feature of the $i$-th sketch sample belonging to the class $y_i$. $W\in \mathbb R^{d\times C}$ denotes the weights of all $C$ class centres with $B\in \mathbb R^C$ as bias terms. We transform $W_{y_j}^T f(x_i)$ to $\parallel W_{y_j}\parallel \parallel f(x_i)\parallel \cos \theta_{i,y_j}$ where $\theta_{i,y_j}$ is the angle (\ie cosine distance) between $f(x_i)$ and $W_{y_j}$. For ease of \doublecheck{analysis}, we further eliminate the bias term and set $\parallel W_{y_j}\parallel$ to $1$. This gives us a modified Softmax formulation as follows:
\begin{equation}
\label{eq:softmax_norm}
\hspace{-1mm}
\widetilde L_{sm}(x_i) = -\log\frac{e^{\parallel f(x_i)\parallel\cos \theta_{i,y_i}}}{e^{\parallel f(x_i)\parallel\cos \theta_{i,y_i}}+\sum\limits_{j=1, j\neq i}^C e^{ \parallel f(x_i)\parallel\cos \theta_{i,y_j}}} 
\end{equation}
Assume that each class has the same number of samples and that all samples are well-separated, we could obtain the lower bound of $\widetilde L_{sm}$ as (details in the supplementary):
\begin{equation}
\label{eq:softmax_lower_bound}
\widetilde L_{sm} \geq \log(1+(C-1)e^{-\frac{C}{C-1}\parallel f(x_i)\parallel}))
\end{equation}

\noindent Astute readers may already notice the catastrophic implication under the loss function $\widetilde L_{sm}$: the optimisation process can be dominated towards maximising $\parallel f(x_i)\parallel$ and completely independent of $\theta$ at its worst, derailing from the very goal of categorisation. A naive solution is to unit normalise $\parallel f(x_i)\parallel$ so that $\widetilde L_{sm}$ would focus on optimising $\theta$ again. This however introduces another side effect. To see it clearer, imagine the extreme ideal case where a sample is embedded infinitesimally close to its centre. As such, the gradient of $\widetilde L_{sm}(x_i)$ w.r.t the ground-truth label $y_i$ is $1-\frac{e^1}{e^1+(C-1)e^{-1}}$ (0.931 when C=100, 0.993 when C=1000), which means that the model will undesirably back-propagate large gradients even when samples are well separated. To get around this seemingly opposing role of feature magnitude, a similar compromise is often undertaken, where $\parallel f(x_i)\parallel$ is first cancelled out from the formulation (\ie  $\parallel f(x_i)\parallel =1$) and prefixed with a global scalar $s$ to simulate its critical effect for numerical stability under cross-entropy loss. We are now ready to write down a normalised Softmax form as enjoyed by many existing works in the literature \citep{normface,sphereface,cosface,arcface}:

\begin{equation}
\label{eq:nsoftmax}
L_{norm\_sm} = -\log\frac{e^{s\cos\theta_{i,y_i}}}{e^{s\cos\theta_{i,y_i}}+\sum\limits_{j=1, j\neq i}^C e^{s \cos\theta_{i,y_j}}} 
\end{equation}
\noindent where the exact value of $s$ is empirically set.

\keypoint{A new perspective on $\parallel f(\cdot)\parallel$.} One problem with Eq.~\ref{eq:nsoftmax} is its inclination to treat every sketch sample equally recognisable -- all $\cos\theta_{i,y_i}$s are optimised towards the same optimum of being as close to the class centre as possible. This loss of instance discrimination comes against how we perceive human sketch data in practice, where people can draw dramatically different bunnies while retaining recognisability (Fig.~\ref{fig:intro}(a)). These bunnies are certainly not of equal quality, and nor should their feature distances be the same to the class centre. A natural question to ask is then that instead of over-simplifying the role of feature magnitude $\parallel f(\cdot)\parallel$ to a constant scalar, can we exploit it to encourage the \textit{establishment of quality semantics within the same 
class} so that $\cos\theta_{i,y_i}>\cos\theta_{j,y_i}$ when $x_i$ is of significantly better quality than $x_j$? We give an affirmative answer to this question. We show by carefully tuning the interplay between $\parallel f(\cdot)\parallel$ and $\cos\theta$ into a unified framework (Sec.~\ref{sec:gacl}), $\parallel f(\cdot)\parallel$ induces an instance-discriminative feature space geometry that permits quality discovery.

\subsection{Geometry-Aware Classification Layer}
\label{sec:gacl}
Our goal is to inject feature magnitude $\parallel f(x_i)\parallel$ as a learnable variable into Eq.~\ref{eq:nsoftmax} and make it work adaptively with $\cos\theta_{i,y_i}$. For that, we introduce a new formulation upon Eq.~\ref{eq:nsoftmax} by replacing $s\cos\theta_{i,y_i}$ with a compound function $A(q_i,\theta_{y_i})$\footnote{For notation simplicity, we use $q_i$ and $\theta_{y_i}$ to represent $\parallel f(x_i)\parallel$ and $\theta_{i,y_i}$ respectively.}. We name it Geometry-Aware Classification Layer (GACL). Denoting $\sum_{j=1, j\neq i}^C e^{s\cos\theta_{i,y_j}}$ as $R$, GACL transforms Eq.~\ref{eq:nsoftmax} to:
\begin{equation}
    \label{eq:GACL_general}
    L_{GACL}(q_i,\theta_{y_i}) = -\log\frac{e^{A(q_i,\theta_{y_i})}}{e^{A(q_i,\theta_{y_i})}+R} 
\end{equation}

\noindent The success of GACL thus relies critically on the design choice of $A(q_i,\theta_{y_i})$, for which we define three necessary constraints for its success.

\begin{itemize}[leftmargin=*]
    \item \textbf{Geometry constraint.} If $q_i$ is a good proxy for quality measurement, it should be larger than $q_j$ when $\theta_{y_i}$ is geometrically lying closer to the class centre than that of $\theta_{y_j}$, \ie  $(q_i-q_j)(\theta_{y_i}-\theta_{y_j})\leq 0$.
    \begin{itemize}[leftmargin=*]
    \item \textit{Condition on $A(q_i,\theta_{y_i})$.} Given two sketches with different value pairs of $(q_i,\theta_{y_i})$ and $(q_j,\theta_{y_j})$, we assume that both have reached optimal recognisability/optimisation equilibrium, $L_{GACL}(q_i,\theta_{y_i})=L_{GACL}(q_j,\theta_{y_j})$. We perform a Taylor expansion of the left-hand side:
    \begin{equation}
    \label{eq:taylor}
    \begin{aligned}
  &L_{GACL}(q_i, \theta_{y_i})  \approx L_{GACL}(q_j, \theta_{y_j}) + \\
  & (q_i - q_j) \bigtriangledown_q L_{GACL} + (\theta_{y_i} - \theta_{y_j})\bigtriangledown_\theta L_{GACL}
\end{aligned}
\end{equation}
\noindent where we have dropped higher order terms. To ensure $(q_i-q_j)(\theta_{y_i}-\theta_{y_j})\leq 0$, it is then easy to obtain the condition to which $A(q_i, \theta_{y_i})$ must satisfy:
\begin{equation}
    \frac{\bigtriangledown_q A(q_i, \theta_{y_i})}{\bigtriangledown_\theta A(q_i, \theta_{y_i})} > 0
\label{eq:geo_constraint}
\end{equation}
\end{itemize}

\item \keypoint{Co-optimisation constraint.} A prerequisite for the geometry constraint to hold is to ensure that $\theta_{y_i}$ can be properly optimised. Unfortunately, Eq.~\ref{eq:softmax_lower_bound} tells us this doesn't come easily as the training dynamics can be completely dominated by $q_i$ and thus become less irrelevant with the optimisation of $\theta_{y_i}$. We alleviate this issue by asking that any updates on $q_i$ won't consequently impede the progress from $\theta_{y_i}$:

\begin{itemize}[leftmargin=*]
\item \textit{Condition on $A(q_i,\theta_{y_i})$.} We use one step of gradient descent to model the effect of the update on $q_i$ to $\theta_{y_i}$:
\begin{equation}
\begin{aligned}
    q_i' &= q_i - \xi \bigtriangledown_q L_{GACL}(q_i, \theta_{y_i}) \\
    \theta_{y_i}' &= \theta_{y_i} - \xi \bigtriangledown_\theta L_{GACL}(q_i, \theta_{y_i})\mid_{q_i=q_i'}
\end{aligned}
\end{equation}

\noindent where $\xi$ is the learning rate. Our goal is then to ensure the non-destructive $\theta_{y_i}$ learning, \ie  $\bigtriangledown_\theta L_{GACL}(q_i, \theta_{y_i})\mid_{q_i=q_i'} \geq 0$. This translates to the constraints on $A(q_i, \theta_{y_i})$ as:

\begin{equation}
    \bigtriangledown_\theta A(q_i, \theta_{y_i})\mid_{q_i=q_i'} \leq 0
\end{equation}

\end{itemize}

\item \keypoint{Optimality constraint.} Assuming the value range of $q_i$ is bounded in $[l_q,u_q]$, we require that $L_{GACL}$ always has an optimal solution $q_i^*$ between $[l_q,u_q]$ in order to prescribe a \textit{valid} quality metric.

\begin{itemize}[leftmargin=*]

\item \textit{Condition on $A(q_i,\theta_{y_i})$.} We assume $L_{GACL}$ is a convex function of $q_i$ (\ie  $\bigtriangledown^2_q L_{GACL}(q_i,\theta_{y_i})\ge 0 \Rightarrow \bigtriangledown^2_q A(q_i,\theta_{y_i})\le 0$), which naturally yields to a global optima. The existence of an optimal solution $q_i^*$ in $[l_q,u_q]$ then translates to the following condition to be held: ${\bigtriangledown_{q} L_{GACL}(l_q, \theta_{y_i}) < 0}$ and ${\bigtriangledown_{q} L_{GACL}(u_q, \theta_{y_i}) > 0}$ (because the first derivative of $L_{GACL}$ to $q_i$ is monotonically non-decreasing). Given 
$\bigtriangledown_{q} L_{GACL} = -\frac{R}{e^{A(q_i,\theta_{y_i})}+R}\bigtriangledown_q A(q_i, \theta_{y_i})$ and $\frac{R}{e^{A(q_i,\theta_{y_i})}+R} > 0$, we obtain our last constraints on $A(q_i,\theta_{y_i})$ by requiring $\bigtriangledown_q  A(l_q, \theta_{y_i}) > 0$ and $\bigtriangledown_q  A(u_q, \theta_{y_i}) < 0$.

\end{itemize}
\end{itemize}

\subsection{GACL Instantiations}
\label{sec:inst}

Aiming towards a thorough inspection, we provide four different types of instantiations\footnote{We apply a linear scaling on $q_i$ in practice to make it work in the proper value range $[l_q,u_q]$, which is omitted here for simplicity.} of $A(q_i,\theta_{y_i})$ with each of which functions on a different conceptual space: (i) \textit{scale}: $A(q_i, \theta_{y_i})=(1-q_i)s\cos\theta_{y_i}$; (ii) \textit{multiplicative angular}: $A(q_i, \theta_{y_i}) = s\cos(q_i\theta_{y_i})$; (iii) \textit{additive angular}: $A(q_i, \theta_{y_i}) = s\cos(\theta_{y_i}+q_i)$; (iv) \textit{cosine}: $A(q_i, \theta_{y_i})=s\cos\theta_{y_i}-q_i$;

It's easy to prove that the first two conditions on $A(q_i, \theta_{y_i})$ are met among the four instantiations (see supplementary for details). The tricky part is the guarantee of the optimality of $q_i$, which is bounded under two specific values \{$l_q,u_q$\} and calls for more efforts to satisfy. Rather than handcrafting the possible values of \{$l_q,u_q$\} in the lens of microscope, we propose a more principled strategy by introducing a score regulariser $G(q_i)$:
\begin{equation}
\label{eq:instant}
    \bigtriangledown_q L_{GACL} = -\frac{R}{e^{A(q_i,\theta_{y_i})}+R}\bigtriangledown_q A(q_i,\theta_{y_i}) + \lambda_g \bigtriangledown_q G(q_i)
\end{equation}

\noindent Since the value of $-\frac{R}{e^{A(q_i,\theta_{y_i})}+R}\bigtriangledown_q A(q_i,\theta_{y_i})$ always remains positive in all instantiations, we simply need to set $\bigtriangledown_q G(u_q)=0$ to meet $\bigtriangledown_q L_{GACL}(u_q, \theta_{y_i}) > 0$. We implement $G(q_i)$ as $\frac{1}{q_i}+\frac{1}{{u_q}^2}q_i$ and then focus on achieving $\bigtriangledown_q L_{GACL}(l_q, \theta_{y_i}) < 0$ for each instantiation scenario in the following discussion.

\begin{itemize}
    \item  $A(q_i, \theta_{y_i})=(1-q_i)s \cos{ \theta_{y_i}}.$ Rewriting Eq.~\ref{eq:instant} gives us:
\end{itemize}
    \begin{equation}
\label{eq:ins_1}
        \bigtriangledown_q L_{GACL} = \frac{Rs\cos\theta_{y_i}}{e^{A(q_i, \theta_{y_i})}+R}+ \lambda_g \bigtriangledown_q G(q_i)
\end{equation}
\noindent We know $0<\frac{R\cos\theta_{y_i}}{e^{A(q_i, \theta_{y_i})}+R}<1$. It is then sufficient to ensure ${\bigtriangledown_q L_{GACL}(l_q, \theta_{y_i}) < 0}$ if $\lambda_g \bigtriangledown_q G(l_q) <-s$. And since $\bigtriangledown_q G(q_i)=-\frac{1}{{q_i}^2}+\frac{1}{{u_q}^2}$, we conclude by requiring $\lambda_g>\frac{-s{l_q}^2{u_q}^2}{{l_q}^2-{u_q}^2}$. 
We set $l_q=0.1,u_q=0.3,s=64$ in our implementation.

\begin{itemize}
    \item \keypoint{$A(q_i, \theta_{y_i}) = s\cos(q_i\theta_{y_i})$}. Rewriting Eq.~\ref{eq:instant} gives us:
\end{itemize}
\begin{equation}
\label{eq:ins_2}
        \bigtriangledown_q L_{GACL} = \frac{Rs\theta_{y_i}}{e^{A(q_i, \theta_{y_i})}+R}\sin(q_i\theta_{y_i}) + \lambda_g \bigtriangledown_q G(q_i)
\end{equation}
\noindent We conduct a similar analysis as above, where given $0<\frac{R\sin(q_i\theta_{y_i})}{e^{A(q_i, \theta_{y_i})}+R}<1$, $0\le\theta_{y_i}\le\frac{\pi}{2}$, we enforce $\lambda_g  \bigtriangledown_q G(l_q) < -\frac{s\pi}{2} \Rightarrow \lambda_g>\frac{-s\pi {l_q}^2{u_q}^2}{2({l_q}^2-{u_q}^2)}$ to meet ${\bigtriangledown_q L_{GACL}(l_q, \theta_{y_i}) < 0}$. 
We set $l_q=1.1,u_q=1.25, s=64$ in our implementation.

\begin{itemize}
    \item  \keypoint{$A(q_i, \theta_{y_i}) = s\cos(\theta_{y_i}+q_i)$}. Rewriting Eq.~\ref{eq:instant} gives us:
\end{itemize}
\begin{equation}
 \label{eq:ins_3}
        \bigtriangledown_q L_{GACL} = \frac{Rs}{e^{A(q_i, \theta_{y_i})}+R}\sin(\theta_{y_i}+q_i) + \lambda_g \bigtriangledown_q G(q_i)
\end{equation}
\noindent Similarly, we require $\lambda_g \bigtriangledown_q G(l_q)<-s \Rightarrow \lambda_g>\frac{-s {l_q}^2{u_q}^2}{{l_q}^2-{u_q}^2}$ to meet ${\bigtriangledown_q L_{GACL}(l_q, \theta_{y_i}) < 0}$.
We set $l_q=0.45,u_q=0.65, s=64$ in our implementation.

\begin{itemize}
    \item \noindent $A(q_i, \theta_{y_i}) = s\cos{ \theta_{y_i}-q_i}.$ Rewriting Eq.~\ref{eq:instant} gives us: 
\end{itemize}

\begin{equation}
 \label{eq:ins_4}
        \bigtriangledown_q L_{GACL} = \frac{R}{e^{A(q_i, \theta_{y_i})}+R} + \lambda_g \bigtriangledown_q G(q_i)
\end{equation}

\noindent Similarly, we require $\lambda_g \bigtriangledown_q G(l_q) <-1 \Rightarrow \lambda_g>\frac{-{l_q}^2{u_q}^2}{{l_q}^2-{u_q}^2}$ to meet ${\bigtriangledown_q L_{GACL}(l_q, \theta_{y_i}) < 0}$. 
We set $l_q=0.35,u_q=0.8, s=64$ in our implementation.

\begin{figure}[t]
    \centering
    \includegraphics[width=\linewidth]{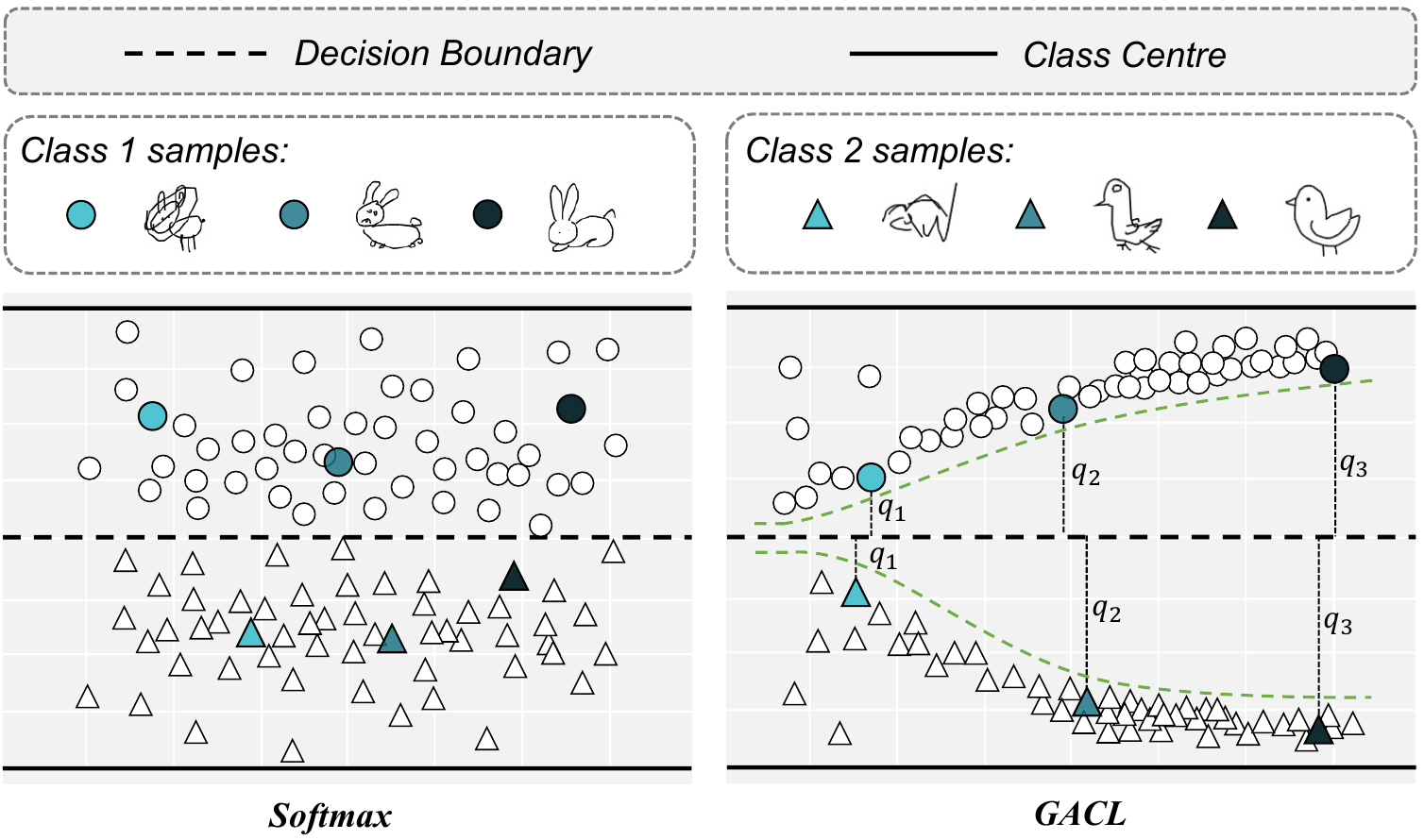}
    \caption{\textit{Geometrical interpretation of $q_i$ under GACL.} Under mild approximation, we show $q_i$ is de facto the distance to the class decision boundary, helping to form a well-structured within-class feature distribution geometry for quality discovery (Example in black bullet point indicates better quality, which is also closer to the respective class centre).}
    \label{fig:geointerp}
\end{figure}

\begin{figure*}[t]
    \centering
    \includegraphics[width=\linewidth]{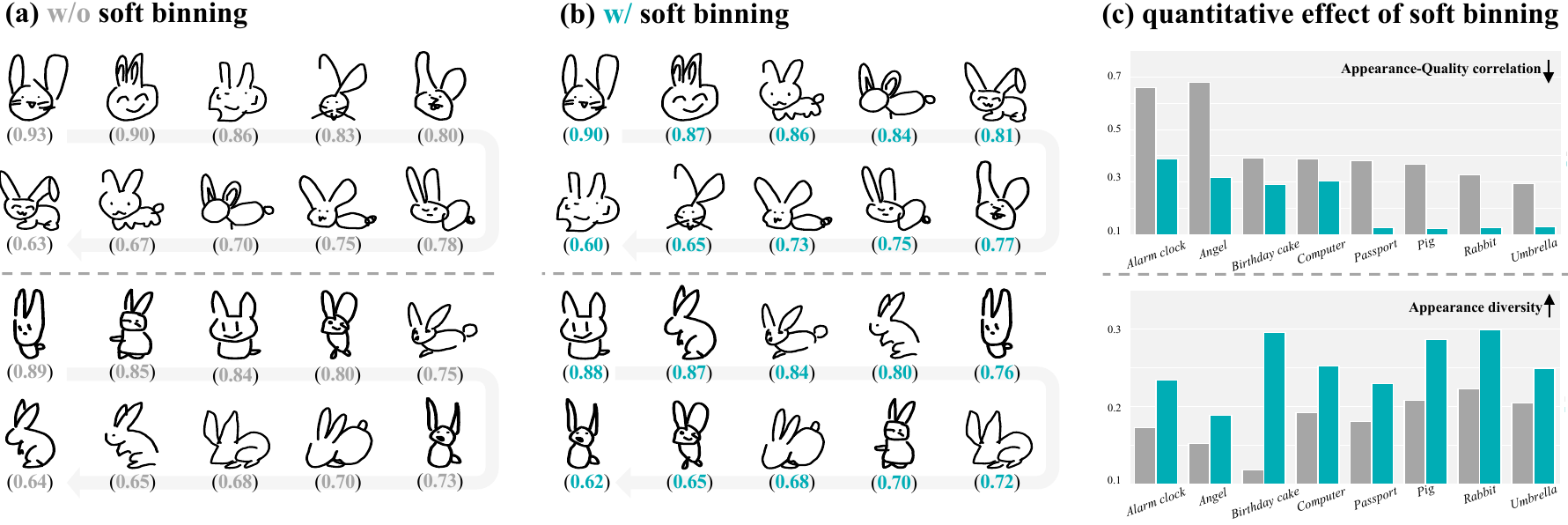}
    \caption{{\textit{Introducing soft binning to reduce bias in quality measurements.} (a) Learning quality scorer $q_i$ with objective Eq.~\ref{eq:GACL_general} can blindly favour certain sketch distributions (frontal-view bunnies in this case) regardless of their visual quality. (b) We alleviate this problem by replacing $q_i$ in Eq.~\ref{eq:GACL_general} to $\hat{q}_i$ in Eq.~\ref{eq:softbin}. $\hat{q}_i$ improves upon $q_i$ with an efficient soft binning technique, which does not involve any new parameter learning. (c) We quantitatively evaluate the effect of soft binning with two perceptual metrics of appearance-quality correlation and appearance diversity. More details in text.}}
    \label{fig:semantic}
\end{figure*}

\subsection{Demystifying $q_i$ as Quality Metric}
\label{sec:explain}
In this section, we provide a different perspective to the role of $q_i$ that comes with a nice geometrical interpretation: Under mild approximation, $q_i$ is the feature distance to class decision boundary, echoing well as a dual task with $\theta_{y_i}$ (\ie co-optimisation), which is the feature distance to class centre. Quality discrimination is then encoded in $q_i$ from such geometrical semantic establishment, as illustrated in Fig.~\ref{fig:geointerp}. To see clearly how $q_i$ represents the distance to the decision boundary, we first review how Softmax is derived as a classification objective. A general formulation to classify an instance $x_i$ among $C$ classes is:
\begin{equation}
\label{eq:hardmax}
\max(\underset{j\neq i}{\max}\{\cos\theta_{y_j}\}-\cos\theta_{y_i},0) 
\end{equation}

\noindent which is the raw ``hardmax'' implying that the target logit score should be greater than the rest. By smoothing the two $\max$ functions with mathematical approximations\footnote{(i) $\LogSumExp(x)$ for $\max(x)$; (ii) $\SoftPlus(x)$ for $\max(x,0)$.}, we arrive at the normalised Softmax in Eq.~\ref{eq:nsoftmax}. The problem with Eq.~\ref{eq:hardmax} is that it completely disregards the within-class feature distribution, where samples are treated equally so long as they belong to the same class label, and thus undermines potential quality discovery. It is then intuitive that we embed an instance-adaptive margin $m_i$ into Eq.~\ref{eq:hardmax} hoping samples with better quality should be pulled farther away from the decision boundary and pushed close to the class centre:

\begin{equation}
\label{eq:adaptivemax}
\max(\underset{j\neq i}{\max}\{\cos\theta_{y_j}\}-\cos\theta_{y_i}+m_i,0)
\end{equation}

\noindent Similarly, by replacing the two $\max$ functions with their soft approximations, we obtain the soft version of Eq.~\ref{eq:adaptivemax}:
\begin{equation}
\label{eq:margin}
\begin{aligned}
    &\log(1 + e^{\log (\sum_{j=1, j\neq i}^C e^{\cos\theta_{y_j}})-\cos\theta_{y_i}+m_i}) \approx \\
    &-\log\frac{e^{{\cos\theta_{y_i}}-m_i}}{e^{{\cos\theta_{y_i}}-m_i} + \sum_{j=1, j\neq i}^C e^{\cos\theta_{y_j}}}
\end{aligned}    
\end{equation}

\noindent Now we can see that apart from global normalisation term $s$, Eq.~\ref{eq:margin} is exactly our GACL framework with $A(q_i, \theta_{y_i})$ instantiated as $s\cos\theta_{y_i}-q_i$, where feature magnitude $q_i$ becomes $m_i$ and hence completes our proof. We omit the elaborations for other three $A(q_i, \theta_{y_i})$ instantiations and believe the discussions above can serve our purpose in providing intuitions on why GACL permits quality discovery: the synergistic interplay between $q_i$ and $\theta_{y_i}$ yields the feature space geometry that importantly gives rise to the notion of quality.

\subsection{Reducing Quality Bias of $q_i$}

{So far, we suggest that GACL originates from the \doublecheck{quality-recognisability} semantic duality, where a value increase on the former would correspond to the greater confidence of the latter. There is however a prerequisite that GACL has been secretly assuming and which we fail to spell out explicitly: GACL only works when the underlying model state has evolved to a point\footnote{This aligns with the generally acknowledged discovery that deep model learning tends to learn easy function first \citep{ hu2020surprising, kalimeris2019sgd, baldock2021deep}, inter-class recognition in our case.} with an adequate level of inter-class separation in order to focus on optimising intra-class feature geometry. To see this clearly, suppose at the early phase of training when GACL struggles to differentiate between the categories of sketch objects. The focus of $\theta_i$ (in Eq.~\ref{eq:taylor}) is inevitably on getting as \doublecheck{close} as possible to its class centre to achieve discriminability. And since $q_i$ \doublecheck{positively correlates with} $\cos\theta_i$, it will be co-optimised to a larger value as well. Question then arises as $q_i$ deviates from its very role of examining visual quality during the said process.}

{Not taking care of such a hidden prerequisite of GACL turns out to be troublesome. We empirically observe a scenario of systematic quality scoring bias from $q_i$ as exemplified in Fig.~\ref{fig:semantic}(a). $q_i$ can favour sketches conforming to particular sub-class patterns by always assigning every sketch in that cluster with higher scores regardless of their de facto visual quality -- e.g., frontal view bunnies are always better than those of side view. Such undesirable phenomenon reaffirms our worry that $q_i$ at the early GACL training stage is encouraged to form tight clusters largely driven by discriminability learning (e.g., frontal-bunnies have more discriminative features such as long ears, etc.), and this is achieved without having quality scoring in mind. Whether such a biased quality understanding accumulated into $q_i$ can be reversed later is purely stochastic. In our case, if bunnies are not from dramatically distinct quality levels, correcting this bias at later learning stages is infeasible.}

{Our solution has a simple motivation -- now that we know updates on $q_i$ are error-prone because $\theta_i$ is not yet ready for intra-class instance recognisability learning (quality scoring), why not hold back quality learning and resume once things are in the better? A naive way is, of course, a hard switch-off with \texttt{stop\_gradient} operation appending on $q_i$, \doublecheck{i.e., $\bigtriangledown_q A(q,\theta)=0$, which would then violate the geometry constraint in Eq.\ref{eq:geo_constraint}.} We instead exploit soft binning \citep{dougherty1995supervised, yang2018deep}, a differentiable workaround that takes a real scalar as input and produces an index of the bins to which it belongs. By narrowing down the working space of $q_i$ into some pre-defined value buckets (\eg $\{0, 0.25, 0.5, 0.75, 1\}$), different $q_i$s within a bucket are processed to having the same value ($0.55, 0.6\Rightarrow 0.5$) and therefore no longer allowed to receiving discriminating gradient updates within a certain value range. 
As such, we temporarily downplay $q_i$ learning in order to sidestep the noisy gradient impacts from \doublecheck{co-optimisation} with $\theta_i$. Concretely, let us assume a continuous quality score $q_i$, that we want to bin into $n+1$ intervals. This leads to the need of $n$ cut points $\beta = [\beta_1, \beta_2, \dots, \beta_n]$ in a monotonically increasing manner, \ie $\beta_1 < \beta_2 < \dots < \beta_n$. Unlike \citep{yang2018deep} that set $\beta$ as trainable variables for learning stochastic neural decision trees, we fix their values by equally splitting between $q_i$'s working range $[l_q, u_q]$. We now construct a one-layer neural network with Softmax as its activation function:
\begin{equation}
\label{eq:softbin}
\begin{aligned}
    o_i &= \text{Softmax}((w q_i + b)/\tau) \;\;\;\; \hat q_i = \sum o_i * \beta
\end{aligned}
\end{equation}
\noindent where $w$ is a constant variable $[1, 2, ..., n+1]$, $b$ is constructed as $[0, -\beta_1, -\beta_1-\beta_2, ..., -\beta_1-\beta_2-\dots-\beta_n]$ and $\tau > 0$ as a temperature hyperparameter. \doublecheck{The idea is to produce an n-length vector that the difference between $x$ and its associated bin value is downplayed and the difference between $x$ and other bin values amplified -- so that we have a vector close to one-shot where the vector entry corresponding to the right bin is 1, i.e., our desirable index\footnote{\doublecheck{It is theoretically verified that when $q_i>\beta_{i}$ and $q_i<\beta_{i+1}$, $o_i$ will likely fall into the interval $(\beta_i, \beta_{i+1})$ in the form of a one-hot categorical encoding as $\tau$ approaches $0$.}}.} Finally, We replace $q_i$ with $\hat q_i$ in Eq. \ref{eq:GACL_general} to obtain our full formulation of GACL. \doublecheck{Note that the whole process is differentiable by design as only Softmax is involved} and we only apply $\hat q_i$ to regularise a specific early period of GACL training. All our quality scoring results presented throughout the paper are still based on $q_i$ that follows a continuous scale.}

{In Fig.~\ref{fig:semantic}(b), we show how GACL, after the involvement of soft binning, is able to perceive visual quality less entangled with the underlying visual semantics (appearance) -- now well-drawn side-view bunnies are able to receive higher quality scores than the badly rendered ones in frontal view. Apart from the qualitative evidence, we also design three discriminative metrics to quantitatively evaluate the efficacy of soft binning as elaborated below.}

\begin{figure}[t]
    \centering
    \includegraphics[width=\linewidth]{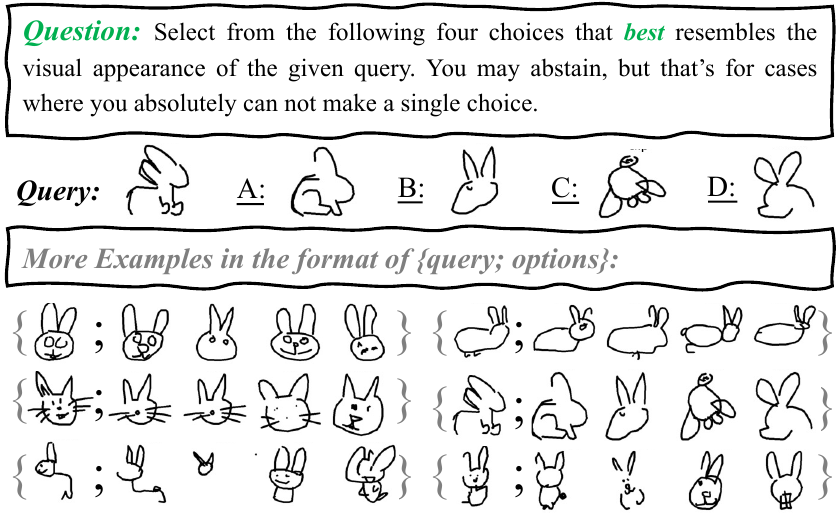}
    \caption{\doublecheck{\textit{Evaluate VGG as sketch appearance encoder.} Given a sketch query and four retrieved sketches by different methods (VGG \citep{simonyan2014very}, d-VAE \citep{rombach2022high}, CLIP \citep{radford2021learning}, SketchClassifier \citep{qu2023sketchxai}), we recruit human participants to decide which of the four sketches presents the best appearance similarity. For illustration purpose, we always adopt the same order here (A for VGG, B for d-VAE...). In practical trials though we always randomise their orders to ensure fair evaluations.}}
    \label{fig:vgg_retrieval}
\end{figure}

\keypoint{Appearance-quality correlation.} {We first aim to examine whether the introduction of soft binning technique is able to decrease the correlation levels between quality scores and instance appearance -- so that bunny pose is becoming a less decisive factor in quality scoring. For that, we randomly sample $500$ sketches from the official test data split of $8$ QuickDraw categories \citep{quickdraw}. We adopt the deep perceptual feature extracted from a VGG network \citep{zhang2018unreasonable} to model the sketch appearance. Pearson Correlation Coefficient (PCC) is then calculated between the pairwise absolute quality difference $X=\mid q_i-q_j\mid$ and their Euclidean feature distance 
$Y=\parallel \text{vgg}(x_i)-\text{vgg}(x_j)\parallel$: $\rho = \frac{cov(X, Y)}{\sigma_X \sigma_Y}$, where $cov$ and $\sigma$ are the covariance and standard deviation function respectively. In the top row of Fig.~\ref{fig:semantic}(c), we can observe that $\rho$ drops significantly under the help of soft binning, in particular for more complex visual categories (Alarm Clock, Angel).}

\keypoint{Appearance diversity.} {Another way to perceive the quality improvements brought by soft binning is to compute the appearance diversity among a group of neighbouring scores -- the higher the diversity, the less functioning bias in $q_i$. Specifically, we take the same data as adopted in PCC setting above and bin the scorings for each category into $10$ uniformly divided value baskets $\{I_1, I_2,..., I_{10}\}$. We then model the appearance diversity within each basket $I_m$ as the variance of pairwise perceptual feature distance:
\begin{equation}
    \varphi_m = \text{mean}(\sum_{i\in I_m} \sum_{j\in I_m, j\neq i} \parallel \text{vgg}(x_i)-\text{vgg}(x_j)\parallel)
\end{equation}
\noindent We report the average variance $\varphi=\frac{1}{10}\sum_{m=1}^{10}\varphi_m$ in the bottom row of Fig.~\ref{fig:semantic}(c) and witness a significant $\varphi$ increase when soft binning of quality scoring is introduced into play.}

\keypoint{Human as judge.} {Finally, from the same test samples, we choose $500$ sketches that are scored most differently with and without the intervention of soft binning. We then recruit $10$ human participants and ask them to make a binary choice for each sketch scoring case: ``which of the scorings is more reflective of the visual quality that better aligns with your perception\doublecheck{.}'' Overall, quality decisions with soft binning are preferred $81.32\%$ of the time. We shall conduct a more comprehensive human study later in Sec.~\ref{sec:humanstudy}.}

\keypoint{\doublecheck{Further note.}} \doublecheck{i) While there have been numerous accounts in the literature that VGG is shown to be a particularly good choice over other mainstream pre-trained deep architectures for sketch-related tasks \citep{yelamarthi2018zero, dutta2019semantically,zhu2020knowledge, wan2021residual,zhu2021learning}, this particular choice might still raise concerns on whether VGG, originally developed for processing coloured images, is really suitable for sketch appearance encoding. To show VGG is a reasonable pick and not necessarily inferior to models that are more amenable to sketch data (task), we design a pilot study that compares VGG with three carefully curated baselines: a visual recognition model specifically tuned on million-scale sketch data \citep{qu2023sketchxai}, a vision-language foundation model that is supposed to have mastered semantic understanding of abundant types of visual data including sketch \citep{radford2021learning}, and a general-purpose visual codebook that is agnostic to data modality due to its pure reconstruction nature and thus inherently sketch-friendly \citep{rombach2022high}. We use sketch-to-sketch retrieval as a task proxy to examine how good a model is in representing sketch appearance. The idea is given a sketch query and a sketch gallery, the better a sketch model the more similar the top-ranked retrieved sketches are. We recruit 5 participants and ask each of them to conduct 200 trials. At each trial, given a sketch and the four choices of different retrieval results by the above-mentioned four different models, we ask a participant to pick \textit{one} choice that best resembles a sketch query appearance considering both global consistency and local similarity (Fig. \ref{fig:vgg_retrieval}). Participants can also choose to abstain if they simply can not figure out a single best option. Overall, VGG is preferred 24.01\% of the time when people believe there is one model standing out amongst the four. The abstention rate is unusually high reaching 22.04\%, providing further evidence that VGG is on par with with other competitors. ii) Results in Fig. \ref{fig:semantic} point to the nuanced nature of different semantic categories, for example, the \{Angel, Alarm Clock\} categories are shown to have less appearance-quality correlation compared with that of \{Pig, Rabbit\}. We provide an explanation here and attribute such phenomenon to the inherently different sketching dynamics human undertakes when purposed for the conceptual interpretation of different object categories. The hypothesis is that for categories like angel or alarm clock, there exists less argument among people on what are the key recognisable sketch traits and how to render them. The resultant sketch quality is then more likely about the aesthetics of \textit{local details}, representing a deeper focus on the drawing skill itself. Rendering pig and rabbit, however, have more holistic structural variations and exhibit greater perceptual diversity. It is thus inevitable that quality scoring here correlates more with the basic structures of visual appearance, despite our effort to reduce such bias via the technique of ``soft binning''. Note that since our human study adopts a comparative setting for evaluating GACL quality scores other than the quest of assigning a real absolute value for each sketch, it is less likely to be affected by the category discrepancy across notions of quality, \eg even the numeric range of quality values for each category varies, the validity of comparative probe remains.
}

\subsection{$q_i$ for Quality-Guided Sketch Generation}
\label{sec:generate_method}

{In this section, we show that $q_i$ can be re-purposed as a plug-and-play quality critic into existing sketch generative models for quality-guided sketch generation -- the method of which produces the results in the Sec.~\ref{sec:generate}. To our best knowledge, either conditional or unconditional sketch generative models \citep{quickdraw,song2018learning,ge2020creative,das2021cloud2curve} are currently quality unattended. Without loss of generality, we choose SketchRNN \citep{quickdraw} as our generative model backbone. SketchRNN takes the form of a variational auto-encoder \citep{kingma2013auto}, with a bidirectional LSTM as encoder that projects a sequence of sketch points $s$ into latent embedding $z=E(s)$, and an LSTM decoder $D(\cdot)$ conditioned on $z$ to reconstruct $s$. We refer the readers to the SketchRNN paper for more details.}

{We portray the problem of quality-guided sketch generation as an iterative process of latent feature discovery. This means given $s$ and its initial latent representation $z_0$, we aim to traverse in the latent space to a target $z$ that is not too far to $z_0$ but with a significantly higher quality score under $q(\cdot)$, which is formulated as:}
\begin{equation}
\begin{aligned}
\label{eq:generation}
    & L_{latent} = (q_{max}-q(D(z))) + \alpha(z-z_0)^2 \\
    & z:= z-\lambda \bigtriangledown_z L_{latent}
\end{aligned}
\end{equation}

\noindent {where $q_{max}$ corresponds to different $u_q$ values under different instantiations, $\alpha$ and $\lambda$ are two hyper-parameters controlling relative importance of identity preservation and gradient descent step size.}

\keypoint{Non-differentiable point sampling.} {Eq.~\ref{eq:generation} requires gradients flowing from $q(\cdot)$ back to $D(\cdot)$, which is problematic in practice as $D(\cdot)$ involves non-differentiable operation during the sampling of Gaussian Mixture Model (GMM)\footnote{Modelling each sketch point as a Gaussian Mixture Model is adopted in most existing sketch generations works \citep{quickdraw,song2018learning,sketchhealer}. This is in contrast to the single-modal normal distribution that corresponds to common $L_2$ regression loss for maximum likelihood estimation.} for sketch point generation. Putting formally, suppose a GMM is instantiated with $M$ normal distributions, the backpropagation discontinuity happens when we need to sample from a categorical vector $\Pi$ of length $M$ that represents the mixture weights. We get around this issue by: (i) Gumbel-Softmax \citep{jang2016categorical}, a differentiable approximate sampling mechanism for categorical variables via reparameterisation trick. (ii) straight-through gradient estimator \citep{bengio2013estimating} for discrete actions in $\argmax$. Combining both turns the once indifferentiable $y=\onehot(\underset{i}{\argmax}(\Pi_{i}))$ to:}
\begin{equation}
\begin{aligned}
\label{eq:sampling}
    & y_{soft}=(\Pi_1',\Pi_2',...\Pi_M') \;\; \\
    & \Pi_i' = \frac{exp((\Pi_i+g_i)/\tau)}{\sum_{j=1}^{M}exp((\Pi_j+g_j)/\tau)} \\
    & y_{hard} =\onehot(\underset{i}{\argmax}(y_{soft})) \\
    & y_{new} = \stopgradient(y_{hard}-y_{soft})+y_{soft}
\end{aligned}
\end{equation}

\noindent {$g_1,g_2,...,g_M$ are i.i.d samples drawn from $\Gumbel(0,1)$\footnote{$Gumbel(0,1)$ is sampled by first drawing $u \sim \Uniform(0,1)$ and computing $g_i=-\log(-\log(u))$.}, $\tau$ is the Softmax temperature that interpolates between discrete one-hot-encoded categorical distributions and continuous categorical densities. By replacing $y$ with $y_{new}$, we can now proceed with Eq.~\ref{eq:generation} in an end-to-end manner.}

\keypoint{\doublecheck{On sketch-based image generation.}} \doublecheck{With a quality-enhanced version of original $s$ denoted as $s_{qe}$, we aim to further demonstrate the positive impact this might bring along to the traditional problem of sketch-based image generation. We settle our investigation under the context of ControlNet \citep{zhang2023adding}, which is a SoTA image generation system adapted from the highly acclaimed StableDiffusion \citep{rombach2022high}. Apart from the conventional text input that partially makes up the style and composition of a visual scene, ControlNet allows an extra form of conditioning control from pixelated visual inputs, such as edgemap (pseudo sketch). We argue that directly feeding ControlNet a raw human sketch $s$ with significant level of line deformation and structural abstraction would produce less appealing results as it was designed for edgemaps perfectly projected from natural images. The natural follow-up is then that with $s_{qe}$, an alternative to $s$ that comes closer to an edgemap structurally should have a play towards better image generation. Specifically, We see both $s$ and $s_{qe}$ as valid essential conditions to ControlNet, where we want the $s_{qe}$ to dominate early diffusion time steps (structure forming phase \citep{wu2023uncovering}) and $s$ to take on more role towards the end (detail rendering phase). In practice, we simply interpolate between the latent representations of $s$ and $s_{qe}$ with a linear function and use a single-valued weighting to adjust their relative importance.
}

\section{Experiments}

\keypoint{Settings.} We evaluate our approach on the largest human free-hand sketch dataset to date, QuickDraw \citep{quickdraw}, which is collected via an online game where the players are asked to sketch a given category name in less than 20 seconds. QuickDraw contains 345 object categories with each containing 70k, 2.5k, 2.5k samples for training, validation and testing respectively. We follow the tradition \citep{sketchformer,sketchaa} of using 7k samples per category for training and all testing data for evaluation (862k sketches in total). We implement $f(\cdot)$ as a two-layer BiLSTM \citep{lstm} with 1024 hidden units, and classification head $W$ with MLPs of dimension 2048-1024-345. Adam \citep{kingma2014adam} optimiser is adopted with initial learning rate 1e-3 and a per-epoch cosine annealing schedule for gradient warm restarts \citep{loshchilov2016sgdr}. We train each individual trial for 20 epochs with a batch size of 256, and pre-process vector sketch data to absolute coordinates normalised within range [0, 1]. \doublecheck{We apply soft binning on $q_i$ with 5 cut points for the first 5 epochs and progressively increase the number of cut points to 20 following a linear scheme. \footnote{\doublecheck{Admittedly without an exhaustive search, we do conduct some ablation on the number of cut points and its impact on quality discovery. We find that setting the right number for the first few epochs matters greatly, with $5$ being a reasonable choice (over $3, 7$). Progressively climbing up to a larger bin number in the later epochs is also shown slightly superior to that of an abrupt change, \ie 5 to 20 without transitions in between.}}} Lastly, we denote our four instantiations of GACL (Sec.~\ref{sec:inst}) as \textbf{Ours-Sca}, \textbf{Ours-Mul}, \textbf{Ours-Add} and \textbf{Ours-Cos} respectively.

\begin{figure*}[t]
    \centering
    \includegraphics[width=\textwidth]{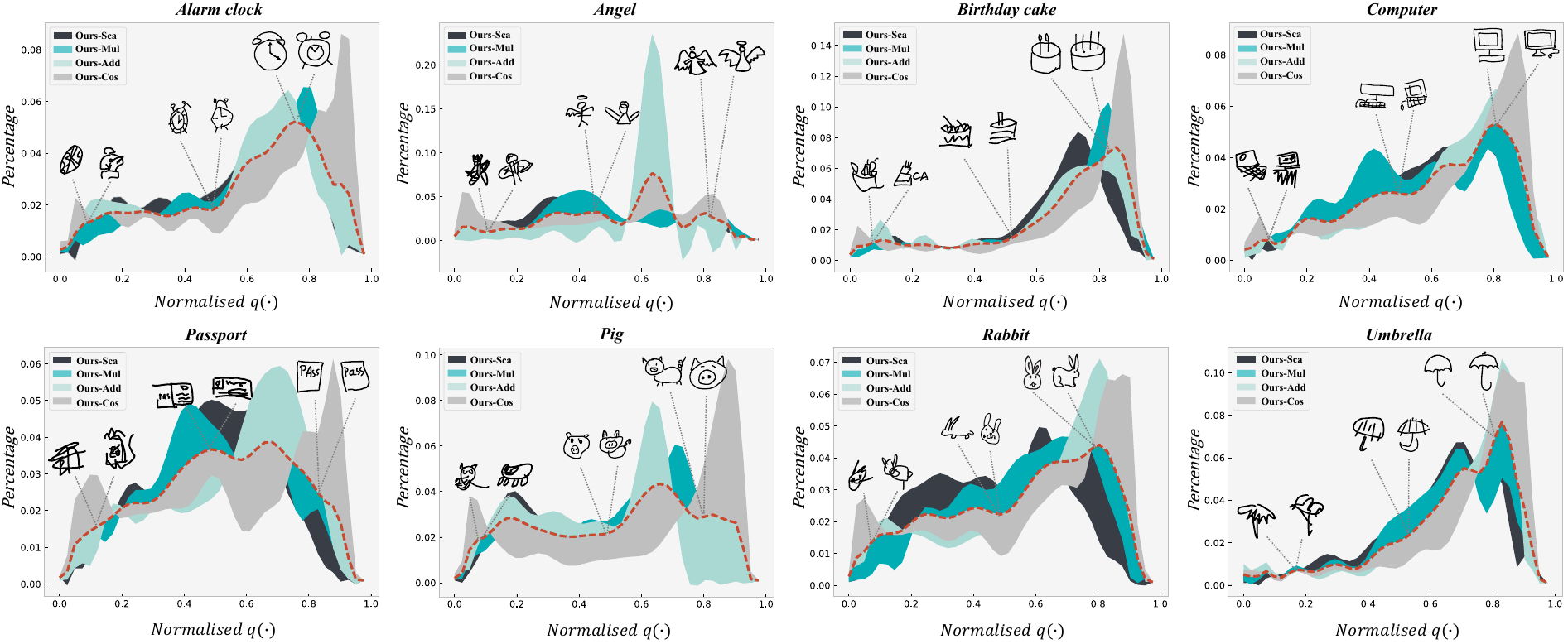}
    \caption{\textit{Qualitative visualisation of $q(\cdot)$ scoring distribution for test samples from different categories.} Selected sketch examples receive similar quality scores under all four GACL instantiations with relative difference less than 0.048. Dashed lines indicate the mean percentage values of sketches falling into a specific scoring range among four instantiations in Sec.~\ref{sec:inst}. Shaded area highlights individual differences. More details in text.}
    \label{fig:visual}
\end{figure*}

\keypoint{Annotators.} \doublecheck{Throughout the experimental session, we have adopted human study either as a quantitative tool to verify our key claims (Sec.~\ref{sec:recog}) or as an auxiliary mean for proof of concept (Sec.~\ref{sec:further}). It is therefore essential that we recruit a pool of annotators whose consent has been obtained and reliability vetted. To identify the annotators that are inclined to produce highly subjective and biased quality scorings, we have opted towards one simple strategy that is commonly adopted in the literature \citep{bell2015learning}: Sentinel and Duplication. Sentinels ensure that bad workers are quickly blocked, and Duplication ensures that the remaining good workers subjected to unstable incompatible decision makings are caught. Specifically, we first crowdsourced 100 workers and asked each of them to conduct 1,000 trials. At each trial, two sketches are given and a binary choice is required for selecting the one with better quality. Sentinels are then the secret test tasks randomly mixed into the existing trials, where workers must agree with the ground truth we supply in advance. In total, we have 40 sentinels and workers who make more than 10 mistakes and are prevented from further submission. Another factor that is also crucial to high quality annotations, especially under a binary choice setting, is whether workers can contribute same answer for the same trials across different points of time -- so that their success can be duplicated and not a matter of luck. We collect two copies of the same task and insert them (40 in total) randomly into the 1,000 trials each worker undertakes. We deem workers who can't reproduce 70\% of their decisions as unqualified. Overall, with these two strategies, we exclude 60 participants from 100 and recruit the best 40 workers to provide the annotation support throughout our empirical evaluations.}

\begin{figure*}[t]
    \centering
    \includegraphics[width=\linewidth]{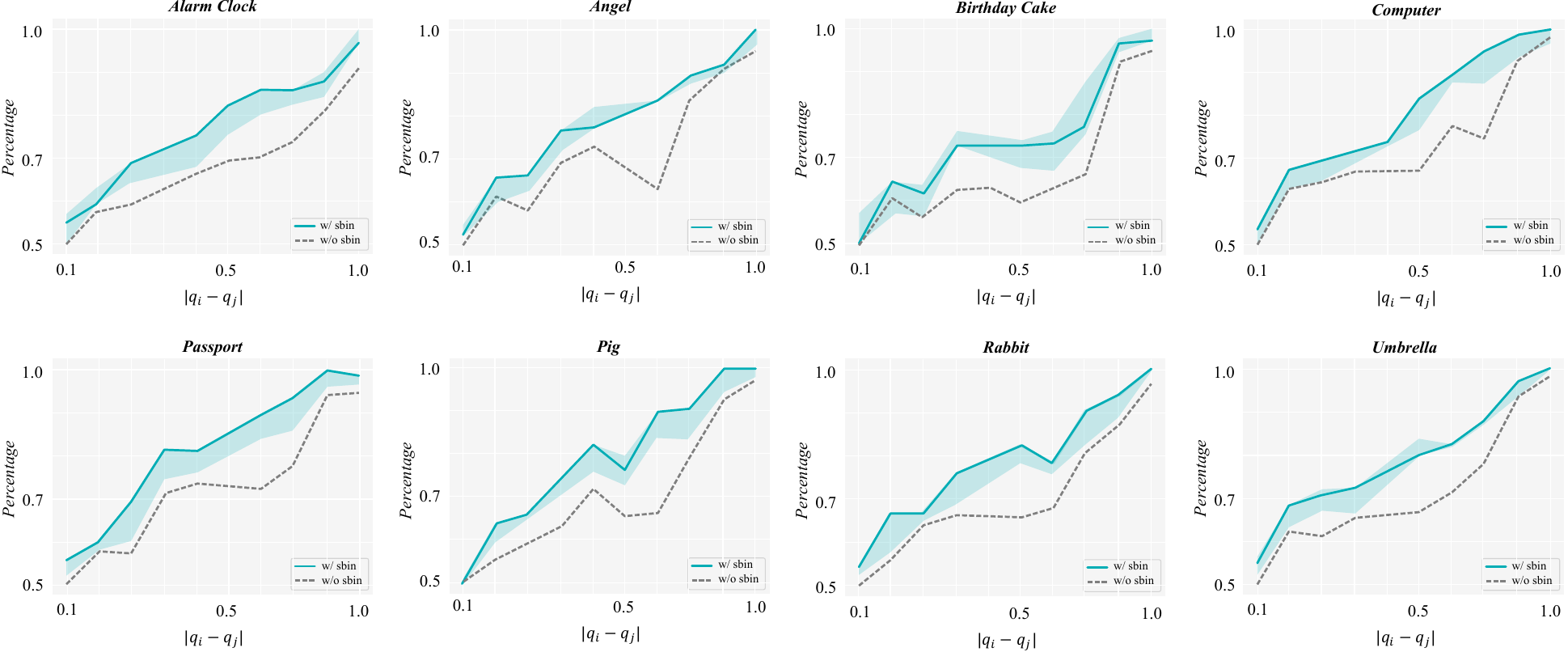}
    \caption{{\textit{Human study results on pairwise quality comparisons across different challenge levels.} In each subplot, we show the performance (percentage of human agreements) of Ours-Cos (solid light green line), Ours-Cos-w/o-sbin (dashed grey line) and the performance deviations to Ours-Cos from other three GACL instantiations defined in Sec.~\ref{sec:inst} (shading area). The smaller value range $\mid q_i-q_j\mid$ resides in, the more challenging the pairwise quality comparison test, the less human consensus with our quality metric is expected.}}
    \label{fig:human}
\end{figure*}

\subsection{GACL Supports Sketch Quality Discovery}
\label{sec:humanstudy}

For empirical evaluation of GACL, we select 8 out of 345 categories in QuickDraw based on the complexity, variety and semantic richness rules outlined in \citep{li2018universal}. In Fig.~\ref{fig:visual}, we first qualitatively visualise their ($500$ per-category random test set samples) distribution of $q(\cdot)$ under different GACL instantiations and demonstrate some exemplary sketch samples separated apart by dramatically different $q$ values. It can be seen that $q(\cdot)$ encodes sketch quality discriminatively in a reasonable way to viewers. Samples corresponding to smaller $q$ values are often aesthetically less pleasing, hard to recognise or simply incomplete and unreliable sketch data. On the other hand, $q(\cdot)$ works from a wide range of perspectives to interpret good sketch quality, including smooth and coherent levels of visual structure rendering (\eg umbrella), local conceptual semantics highlights (\eg passport) and holistic visual aesthetics and richness (\eg angel). It is also understandable that $q(\cdot)$ learned by Ours-Sca/Ours-Mul/Ours-Add/Ours-Cos is noticeably different (shading areas) given the different value domains they are designed to work on. All four GACL instantiations however show similar trend of score distribution change, indicating the possibility of a unified metric depending on the granularity of quality support evaluated, as confirmed in our quantitative evaluation.

\begin{figure}[th]
    \centering
    \includegraphics[width=\linewidth]{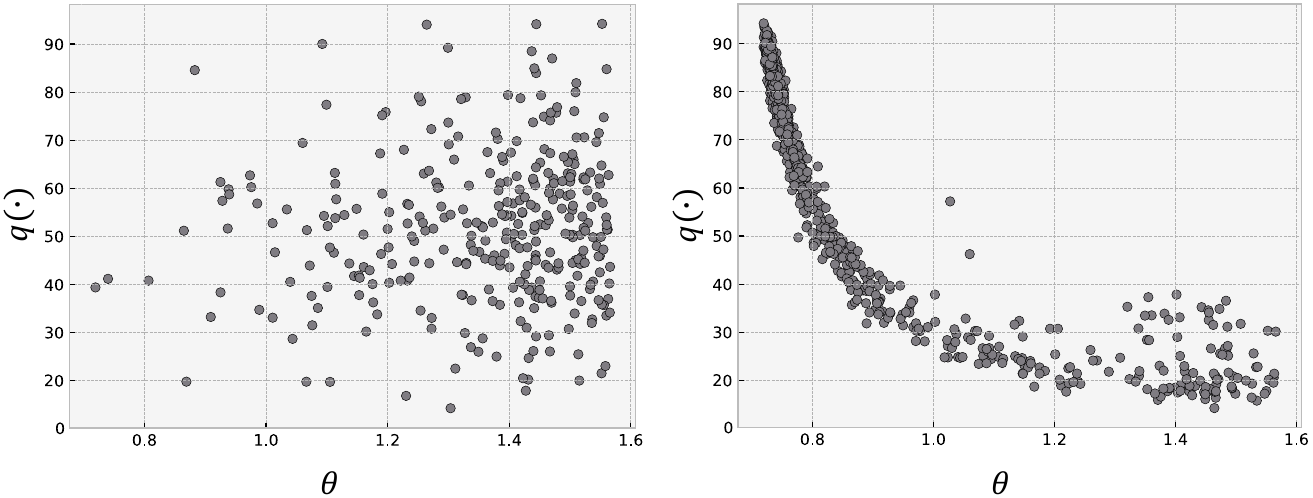}
    \caption{\textit{Interplay between $q(\cdot)$ and $\theta$.} We visualise $q(\cdot)$ and $\theta$ of sketch data using a learned recognition model. Left: Softmax. Right: Ours-Cos. Category: Rabbit.}
    \label{fig:relation}
\end{figure}

\subsection{Quality-Aware Sketch Recognition}
\label{sec:recog}

\begin{figure*}[t]
    \centering
    \includegraphics[width=0.95\linewidth]{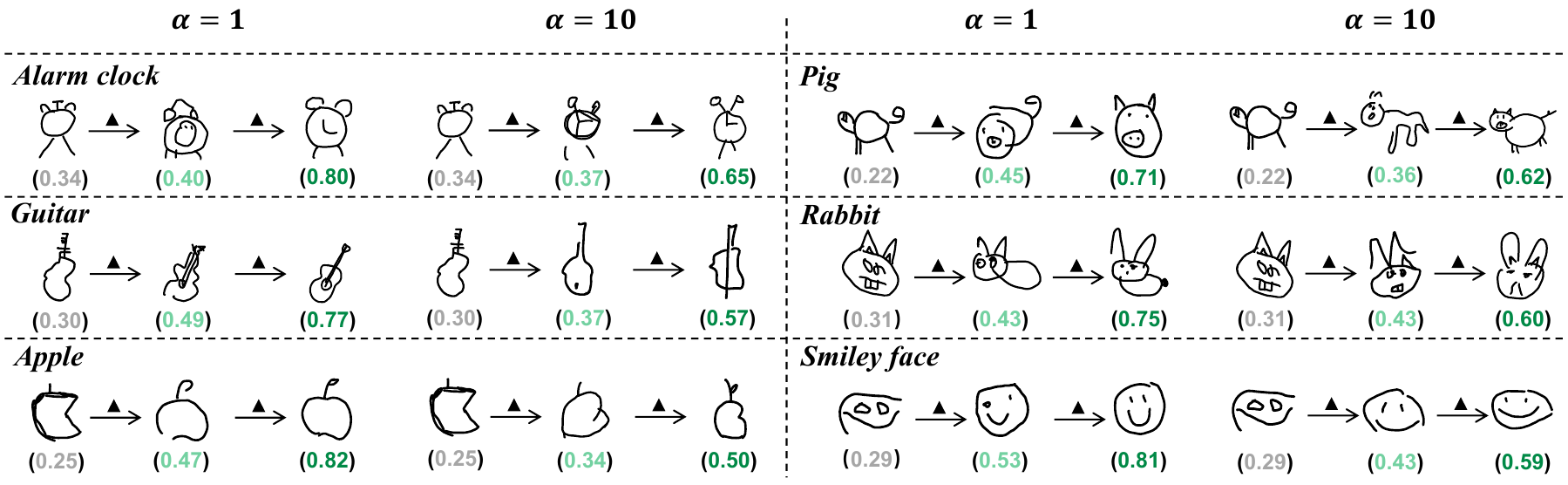}
    \caption{\textit{Quality-guided sketch generation.} Given a sketch input, we show two generation results with better quality (larger $q$ values) separated by $50$ iterations of latent code updates (represented as $\blacktriangle$). $\alpha$ is a hyperparameter that controls self-reconstruction importance. }
    \label{fig:generate}
\end{figure*}
 
It needs more careful inspection when quantitatively evaluating sketch quality. The common way of achieving this by measuring the difference between model predictions and human ground-truth ratings does not apply here as we lack relevant annotations. We further argue that such an approach would be flawed even if we recruit human participants and collect their quality opinions on individual sketches -- it's hard to obtain objective and accurate scores with consensus \doublecheck{given} the subjective and abstract nature of free-hand sketch data. 
{On the other hand, it becomes more feasible for humans to make a comparative choice on the visual quality difference between an image pair as confirmed in the recent literature \citep{fu2015robust, chen2009crowdsourceable, ma2011person}. We then randomly sample two sketches from the test set of each category to form a sketch pair and repeat the sampling process 5,000 times to create $5,000\times 8\times 5=200,000$ pairs in total: number 5 comes from 4 GACL instantiation methods and 1 ablated method of Ours-Cos but \textit{without} soft binning (Ours-Cos-w/o-sbin). 40 participants are recruited with each undertaking 5,000 trials. Each participant is given a sketch pair in each trial and required a binary action on the question ``Select one sketch with better visual quality''. Note that in a separate pilot study, we also offered participants a third option that there is no significant quality difference between the two presented sketches. We however find out that such an option generally encourages the workers to make quicker but less responsible decisions without really trying their best to mine and identify the visual subtleties. We hence remove such an option in our final large scale human study and remain a compulsory binary choice of yes or no.

\begin{table}[t]
    \centering
    \caption{\textit{Quality-aware sketch recognition.} Compare GACL against contemporary sketch recognition baselines on official QuickDraw test split. Numbers are top-1 accuracy.}
    \resizebox{\linewidth}{!}{
    \begin{tabular}{@{}cccc@{}}
    \cmidrule[0.8pt]{1-4}
    -- & -- & \textbf{IJCV'2017}  & \textbf{CVPR'2018}   \\
   \noalign{\smallskip}\cdashline{1-4} \noalign{\smallskip}
    BiLSTM & ResNet-50 & Sketch-a-net & SketchMate   \\
     \noalign{\smallskip}\cdashline{1-4} \noalign{\smallskip}
    79.87\% & 78.76\% & 68.71\% & 79.44\%  \\
    \cmidrule[0.8pt]{1-4}
    \textbf{TOG'2021}  & \textbf{CVPR'2020}  & \textbf{CVPR'2020}  & \textbf{ICCV'2021}   \\
    \noalign{\smallskip}\cdashline{1-4} \noalign{\smallskip}
    SketchGNN & SketchFormer & SketchBert & SketchAA  \\
    \noalign{\smallskip}\cdashline{1-4} \noalign{\smallskip}
    77.31\% & 78.34\% & 80.12\% & 81.51\% \\
    \cmidrule[0.8pt]{1-4}
    \textbf{Ours-Sca} & \textbf{Ours-Mul} & \textbf{Ours-Add} & \textbf{Ours-Cos}  \\
    \noalign{\smallskip}\cdashline{1-4} \noalign{\smallskip}
    81.77\% & 82.02\% & 81.97\% & \textbf{82.52\%} \\ \cmidrule[0.8pt]{1-4}
    \end{tabular}}
    \label{tab:recog}
\end{table}

Among the 160,000 trials, we find on average $78.13\%$ (with standard deviation $0.06\%$) of quality scoring results provided by our four GACL instantiations match human quality preference, validating the efficacy of our proposed quality discovery method. The drop in human agreement percentage from $78.13\%$ (Ours-Cos) to $72.35\%$ (Ours-Cos-w/o-sbin) also confirms the importance of introducing soft binning to ameliorate the otherwise fatal flaw of computational quality bias in GACL. In Fig.~\ref{fig:human}, we further plot separate GACL performance for 8 QuickDraw categories, and in particular how the percentage of human quality consensus changes to the different challenge levels of the pairwise quality comparison. We simulate 10 challenge levels with each level belonging to a unique non-overlapping normalised comparative value range of $\mid q_i-q_j\mid$, \ie [0, 0.1], (0.1, 0.2], ..., (0.9, 1.0]. A smaller value range then should correspond to a more challenging pairwise test since the difference between sketch qualities is more subtle to discern. We divide the results of the $5,000\times 5=25,000$ trails for each category into the 10 said groups and can observe that: (i) Similar to qualitative evidence, human participants can sense notable differences across different GACL instantiations, though they follow similar scoring trends alongside the changes of challenge levels, \eg they all win near-unanimous human agreements in the easiest comparison test of $\mid q_i-q_j\mid \in (0.8, 1.0]$. Ours-Cos appears to provide the best performance (the top enveloping line encircling the shading area) most of the time, and that is why we always adopt it for GACL-empowered applications below unless otherwise mentioned. (ii) The better quality scoring effect resulting from the introduction of soft binning is perceptible to human viewers. The effect is particularly salient for intermediate challenging levels, $\mid q_i-q_j\mid \in [0.5, 0.7]$. This is expected as Ours-Cos-w/o-sbin can already deal with quality comparisons of a gigantic pairwise gap pretty well, which leaves limited room for further improvement. (iii) At the hardest challenge level, $\mid q_i-q_j\mid\in [0, 0.1]$, the performance of GACL is only slightly better than random chances. We argue that this is however not necessarily a limitation of our proposed method. Sketch quality discrimination conducted under such fine-grained scope can be an infeasible task itself.
}

One potential benefit of the proposed GACL framework is to contribute a competitive sketch recognition model as a byproduct -- representation learning discriminates between the quality of sketch instances and thus generalises better by less overfitting on lower quality data as per similar findings in the recent literature \citep{chang2020data,chun2021probabilistic}. To verify, we first visualise the relationship between $q(\cdot)$ and $\theta$ in Fig.~\ref{fig:relation} and confirm that sketch instances of better quality (larger $q$) tend to be more easily recognised (smaller $\theta$) under Ours-Cos, while such phenomenon is failed to be observed in models trained by conventional Softmax loss. We further compare the performance between Ours-Cos and eight contemporary sketch recognition baselines in Tab.~\ref{tab:recog}. 
It can be seen that our approaches achieve \doublecheck{consistent} and \doublecheck{significant} improvements over their no-quality-attended counterpart (\vs BiLSTM) and beat the state-of-the-art sketch recognition method with a noticeable margin (\vs SketchAA).

\begin{figure*}
    \centering
    \includegraphics[width=\linewidth]{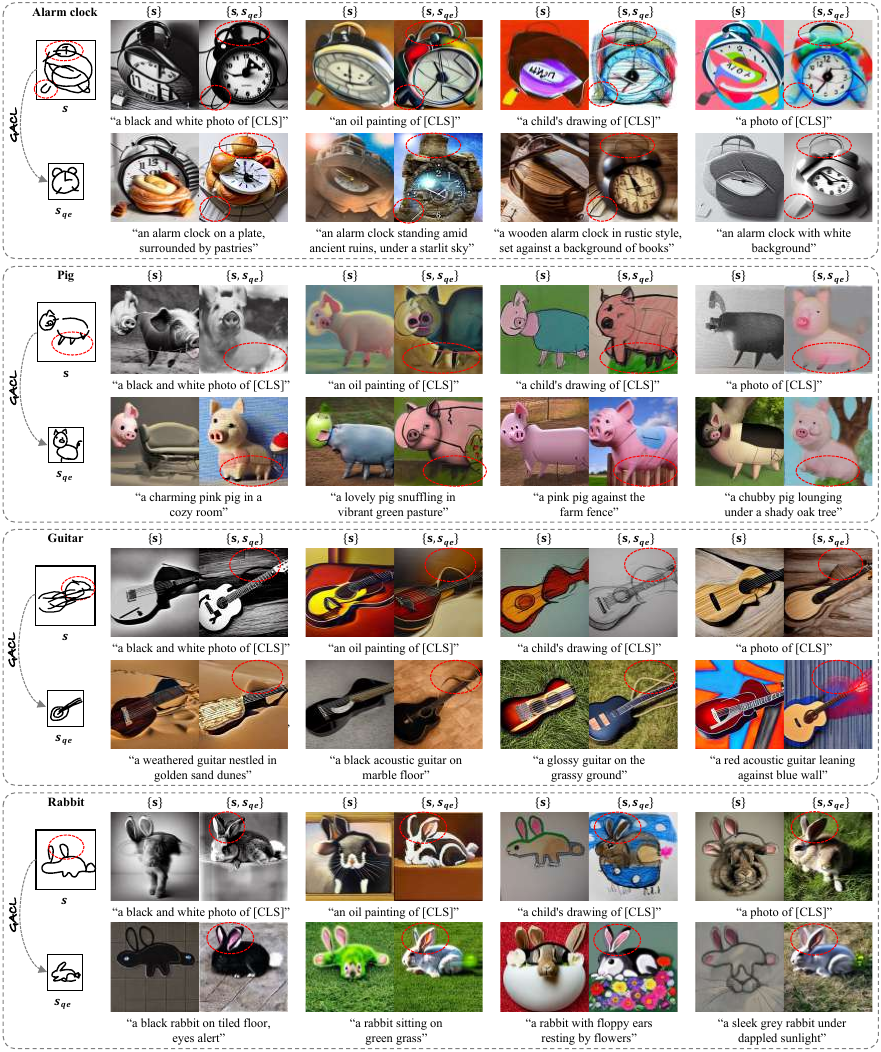}
    \caption{\textit{Impact of GACL on sketch-based image generation.} \doublecheck{We show given a raw human sketch input $s$, how a quality-enhanced sketch $s_{qe}$ enabled by GACL can be leveraged, in a co-play with ControlNet, for better image generation. Other than the prompt contents we articulate above, each is further appended with ``..., best quality, extremely detailed.''. Individual sketch traits are selectively flagged to highlight whether they get preserved in their image counterparts.}}
    \label{fig:controlnet}
\end{figure*}

\subsection{Quality-Guided Sketch Generation}
\label{sec:generate}

In this section, we show that $q(\cdot)$ learned under GACL can be used to guide sketch generative models towards higher quality exploration in a post-hoc iterative manner. A decisive factor affecting the synthesis outcome is the hyperparameter term $\alpha$ that balances the weighting between self-reconstruction and quality improvement -- in our setting, a larger $\alpha$ value prefers the former. We showcase some examples in Fig.~\ref{fig:generate} between generation process under two distinct $\alpha$ values and can observe that (i) our learned quality metric $q(\cdot)$ is indeed a useful drop-in module to enrich existing generative sketch models with a quality dimension. By sliding along the iteration steps, we can customise the extent of quality improvement. (ii) the choice of $\alpha$ matters. While a lower $\alpha$ value generally leads to the generated sketches with higher $q$ values, it can also result in new visual imagery that is completely disconnected from the input (\eg pig and rabbit), failing the quality \textit{guidance} intent. This suggests one future work direction on exploration of an adaptively set $\alpha$ value (\cf fixed) to strike a better balance between quality improvements and identity preservation. \doublecheck{Lastly, we compare the ControlNet generation results before (``\{$s$\}'') and after (``\{$s$, $s_{qe}$\}'') the involvement of GACL in Fig.~\ref{fig:controlnet}. The evidence is compelling where $s_{qe}$ can help to alleviate the structural inconsistency and visual oddities in the images solely enabled by $s$. Interestingly, the intervention of $s_{qe}$ still keeps most of the key visual cues conveyed by $s$ thanks to our step-sensitive approach outlined in Sec.~\ref{sec:generate_method}. The generation process could be otherwise carried away if ControlNet pays attention to $s_{qe}$ for an improper long period of diffusion steps.}

\begin{figure*}[t]
    \centering
    \includegraphics[width=\linewidth]{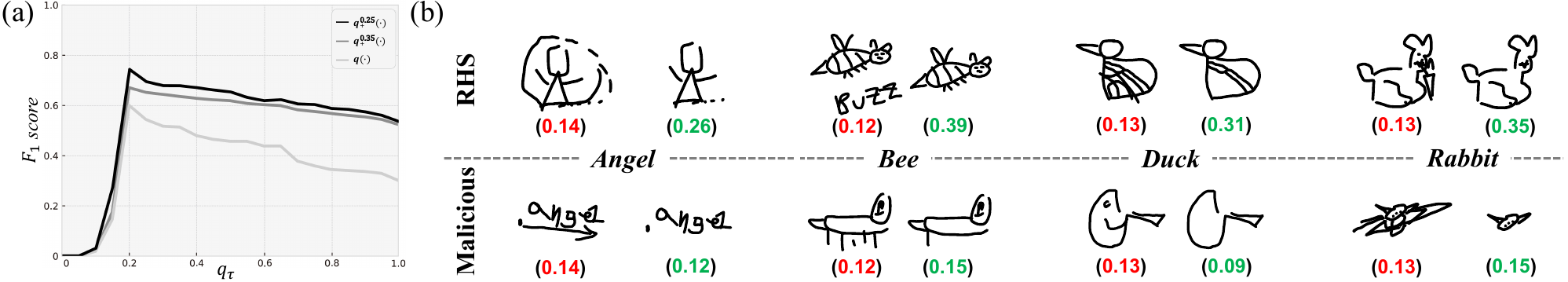}
    \caption{\textit{Quality-enabled sketch attribution.} (a) Comparison between different sketch attributions methods under $F_1$ score. (b) Illustration of malicious and \doublecheck{rare and hard sketch (RHS)} inputs, which both have very low $q$ values (number in red). RHS sketch, however, can uniquely rise to a significantly larger $q$ value (number in green) after some simple stroke removal strategy -- what malicious sketches can't. This gives a chance to attribute between \doublecheck{RHS} and malicious sketch inputs.}
    \label{fig:attr}
\end{figure*}

\subsection{Quality-Enabled Sketch Attribution}

One practical bottleneck for sketch model deployment today is the general lack of method for user sketch attribution -- when the poor model performance is detected, developers can not know whether it comes down to model capacity itself or the malicious\footnote{Random scribbles that do not conform to any semantic concept.} user sketch input. In this section, we aim to examine to what extent our learned $q(\cdot)$ benefits such purpose. Intuitively, if the $q$ value of a sketch is greater than a threshold value $q_\tau$, we deem it as a benign user input (better quality) or otherwise a malicious one. We collect human opinions of their binary decision on whether a given sketch is maliciously intended and form an annotated test set of 2,000 sketches with 1,000 for each sketch type as human ground truth. We adopt $F_1$ score \citep{powers2011evaluation} as our evaluation metric for its ability to balance the performance between precision and recall. The result in Fig.~\ref{fig:attr}(a) shows that with a raw $q(\cdot)$ scorer can achieve the best $F_1$ score of 61.25\% for sketch attribution.

With some additional efforts looking into analysing the attribution disagreements between our method and human annotators, we arrive at one interesting observation: some irregular sketches with excessively long and fragmented strokes or irrelevant personalised decorations deviating from the main rendering objective (Fig.~\ref{fig:attr}(b)) are often treated as non-malicious inputs by human judges, contrary to our model predictions. We term these sketches as \doublecheck{rare and hard sketches (RHS)} and devise a way to prevent our model from attributing them to malicious inputs. The key insight is that despite $q$ values for \doublecheck{RHS} and malicious sketch are all low, the stroke subset of the former can justify a much larger $q$ value -- it does encapsulate a well-recognisable visual object, but just with noisy visual outliers perturbing model predictions. This means given a sketch input and a $q$ value lower than the threshold value $q_\tau$, we can chip in one extra conditioning step before we decide to categorise it into malicious input or not. Specifically, we simply test one stroke at a time and remove it from the input if that can lead to a noticeable increase in $q$ value. We treat a sketch as RHS, \ie non-malicious input, if there exists a partial composition of its strokes that reach a $q$ value more than a pre-set threshold $q_{\max}$ (Fig.~\ref{fig:attr}(b)). We denote such method as $q_+^{q_{\max}}(\cdot)$ and compare with $q(\cdot)$ using two different $q_{\max}$ values in Fig.~\ref{fig:attr}(a). Significant improvements can be observed by taking into account our modelling on RHS sketch input.

\begin{table*}[t]
    \centering
    \caption{\textit{Applying GACL on traditional image quality assessment (IQA) task.} We portray IQA as a classification problem and re-purpose GACL to solve it. We show GACL, originally designed for quantifying sketch data quality, is able to achieve competitive IQA results and new SoTA for four standard benchmarks. Note that the less competitive performance on LIVE and CLIVE does not necessarily indicate a particular disadvantage of GACL. LIVE and CLIVE contain only 799 and 1,162 reference images respectively. Sufficient learning of GACL becomes less feasible with such limited data size. N/A refers to hand-crafted feature modelling, i.e., no deep learning.}
    \resizebox{\linewidth}{!}{
    \begin{tabular}{l||ccccccccccccc}
    \hline
    & Backbone & \multicolumn{2}{c}{LIVE} & \multicolumn{2}{c}{CSIQ }  & \multicolumn{2}{c}{TID2013 } & \multicolumn{2}{c}{KADID } & \multicolumn{2}{c}{KONIQ-10K } & \multicolumn{2}{c}{CLIVE } \\
    \cdashline{2-14}
    & & PLCC & SROCC & PLCC & SROCC & PLCC & SROCC & PLCC & SROCC & PLCC & SROCC & PLCC & SROCC \\
    \hline
    \hline
    DIIVINE \citep{DIIVINE} (TIP'2012)& N/A & 0.908 & 0.892 & 0.776 & 0.804 & 0.567 & 0.643 & 0.435 & 0.413 & 0.558 & 0.546 & 0.591 & 0.588 \\
    BRISQUE \citep{BRISQUE} (TIP'2012)& N/A &  0.944 & 0.929 & 0.748 & 0.812 & 0.571 & 0.626 & 0.567 & 0.528 & 0.685 & 0.681 & 0.629 & 0.629 \\
    ILNIQE \citep{zhang2015feature} (TIP'2015) & N/A & 0.906 & 0.902 & 0.865 & 0.822 & 0.648 & 0.521 & 0.558 & 0.534 & 0.537 & 0.523 & 0.508 & 0.508 \\
    BIECON \citep{kim2016fully} (J-STSP'2016)& Conv$*$2 & 0.961 & 0.958 & 0.823 & 0.815 & 0.762 & 0.717 & 0.648 & 0.623 & 0.654 & 0.651 & 0.613 & 0.613 \\
    HFD \citep{HFD} (ICCVW'2017) & ResNet50 & 0.971 & 0.951 & 0.890 & 0.842 & 0.681 & 0.764 & - & - & - & - & - & - \\
    MEON \citep{ma2017end} (TIP'2017) & Conv$*$4 & 0.955 & 0.951 & 0.864 & 0.852 & 0.824 & 0.808 & 0.691 & 0.604 & 0.628 & 0.611 & 0.710 & 0.697  \\
    WaDIQaM \citep{bosse2017deep} (TIP'2017) & Conv$*$10 & 0.955 & 0.960 & 0.844 & 0.852 & 0.855 & 0.835 & 0.752 & 0.739 & 0.807 & 0.804 & 0.671 & 0.682 \\
    RankIQA \citep{rankiqa} (ICCV'2017) & VGG16 & 0.973 & \textbf{0.974} & 0.912 & 0.892 & 0.793 & 0.780 & - & - & - & - & 0.675 & 0.641 \\
    PQR \citep{PQR} (ICIP'2018) & ResNet50 & 0.971 & 0.965 & 0.901 & 0.873 & 0.864 & 0.849 & - & - & - & - & 0.836 & 0.808 \\
    DBCNN \citep{zhang2018blind} (TCSVT'2018) & VGG16 & 0.971 & 0.968 & \textbf{0.959} & \textbf{0.946} & 0.865 & 0.816 & 0.856 & 0.851 & 0.884 & 0.875 & 0.869 & \textbf{0.869} \\
    TS-CNN \citep{8550783} (TIP'2019) & Conv$*$10 & \textbf{0.978} & 0.969 & 0.904 & 0.892 & 0.824 & 0.783 & 0.701 & 0.680 & 0.724 & 0.713 & 0.688 & 0.669  \\
    CaHDC \citep{wu2020end} (TIP'2020) & Conv$*$10 & 0.964 & 0.965 & 0.914 & 0.903 & 0.878 & 0.862 & 0.804 & 0.811 & 0.840 & 0.825 & 0.744 & 0.738 \\
    MetaIQA \citep{zhu2020metaiqa} (CVPR'2020) & ResNet18 & 0.959 & 0.960 & 0.908 & 0.899 & 0.868 & 0.856 & 0.775 & 0.762 & 0.856 & 0.887 & 0.802 & 0.835 \\
    P2P-BM \citep{ying2020patches} (CVPR'2020) & ResNet18 & 0.958 & 0.959 & 0.902 & 0.899 & 0.856 & 0.862 & 0.849 & 0.840 & 0.885 & 0.872 & 0.842 & 0.844 \\
    HyperIQA \citep{su2020blindly} (CVPR'2020) & ResNet50 & 0.966 & 0.962 & 0.942 & 0.923 & 0.858 & 0.840 & 0.845 & 0.852 & 0.917 & 0.906 & 0.882 & 0.859 \\
    TIQA \citep{you2021transformer} (ICIP'2021) & ViT & 0.965 & 0.949 & 0.838 & 0.825 & 0.858 & 0.846 & 0.855 & 0.850 & 0.903 & 0.892 & 0.861 & 0.845 \\
    UNIQUE \citep{zhang2021uncertainty} (TIP'2021) & ResNet34 & 0.968 & 0.969 & 0.927 & 0.902 & - & - & \textbf{0.876} & \textbf{0.878} & 0.901 & 0.896 & \textbf{0.890} & 0.854 \\
    TReS \citep{golestaneh2022no} (WACV'2022) & ResNet50+ViT & 0.968 & 0.969 & 0.942 & 0.922 & \textbf{0.883} & \textbf{0.863} & 0.858 & 0.859 & \textbf{0.928} & \textbf{0.915} & 0.877 & 0.846 \\
    \cdashline{1-14}
    \multicolumn{1}{l||}{GACL$^{50}$} & ResNet50 & 0.936 & 0.928 & 0.951 &  0.940 & \textbf{\emph{0.946}} & \textbf{\emph{0.940}} & \textbf{\emph{0.898}} & \textbf{\emph{0.884}} & 0.935 & 0.922 & 0.876 & 0.857 \\
    \multicolumn{1}{l||}{GACL$^{100}$} & ResNet50 & \textbf{\emph{0.966}} & \textbf{\emph{0.958}} & \textbf{\emph{0.961}} & \textbf{\emph{0.956}} & 0.936 & 0.927 & 0.884 & 0.880 & 0.936 & 0.922 & 0.862 & 0.858 \\
    \multicolumn{1}{l||}{GACL$^{200}$} & ResNet50 & 0.962 & 0.957 & 0.948 & 0.947 & 0.928 & 0.914 & 0.881 & 0.879 & \textbf{\emph{0.938}}  & \textbf{\emph{0.926}} & \textbf{\emph{0.878}} & \textbf{\emph{0.866}} \\

    \hline
    \end{tabular}}
    
    \label{tab:iqa}
\end{table*}

\section{{GACL for Image Quality Assessment}}

{This paper for the first time introduces the task of evaluating the quality of human sketches. Image quality assessment (IQA), on the other hand, has been an established research topic in the computer vision and image processing field (see recent survey \citep{zhang2021fine}). We aim to show that GACL works for traditional IQA as well and examine that by experimenting on six IQA benchmarks and comparing with eighteen competitive existing methods. Note that unlike our unsupervised sketch quality discovery, these IQA benchmarks come with annotations of human quality scores and thus make the empirical evaluation much more accessible. \doublecheck{For fair comparisons, we select ten different seeds to split the datasets randomly into train/test set (80\%/20\%) and report the median PLCC and SROCC values among the ten different runs -- two most commonly adopted metric for evaluating IQA task in the literature}. In the case of synthetically distorted datasets, the split is ensured to avoid reference image overlap.}

{GACL is easily adaptable to IQA. By discretising the human quality ratings into a histogram with $N$ bins and representing ratings within the same bin as one unified index value, we succeed at transforming a continuous value regression problem to that of a classification one. The granularity of the GACL predicted quality rating is however dependent on the setting of $N$. In Tab.~\ref{tab:iqa}, we experiment with three different $N$ values (50, 100, 200) and observe that GACL can achieve competitive performance on both synthetic (LIVE, CSIQ, TID2013, KADID) and authentic (KONIQ-10K, CLIVE) IQA benchmarks \doublecheck{and crucially not does so with a more power backbone, e.g., compared with TReS, our GACL instantiated with ResNet50 is 25.83\% and 50.72\% fewer in training parameters and flops respectively.}. In particular, we obtain new SoTA with significant improvements over the previously best-reported result, when the underlying dataset (TID2013, KADID, KONIQ-10K) has a sufficient size of more than a few thousand to support GACL training. Why does a classification-based approach such as GACL work for IQA, when the field has firmly suggested that the problem is not best tackled with a classification paradigm \citep{talebi2018nima}\citep{kao2015visual}\citep{wu2011learning}, is however interesting. To answer this, we need to first figure out what makes IQA challenging -- human quality scorings leveraged as ground truth are subjective and noisy. It is then the case that for the same image instance, a classification network will be forced to fit quality ratings for different classes thus making it an inherently ill-posed solution. GACL alleviates the said issue by treating those less reliable human quality ratings as low quality and prevents models from overfitting, \ie model now can digest subjective quality ratings with adaptive confidence.}

\begin{figure*}[t]
    \centering
    \includegraphics[width=\linewidth]{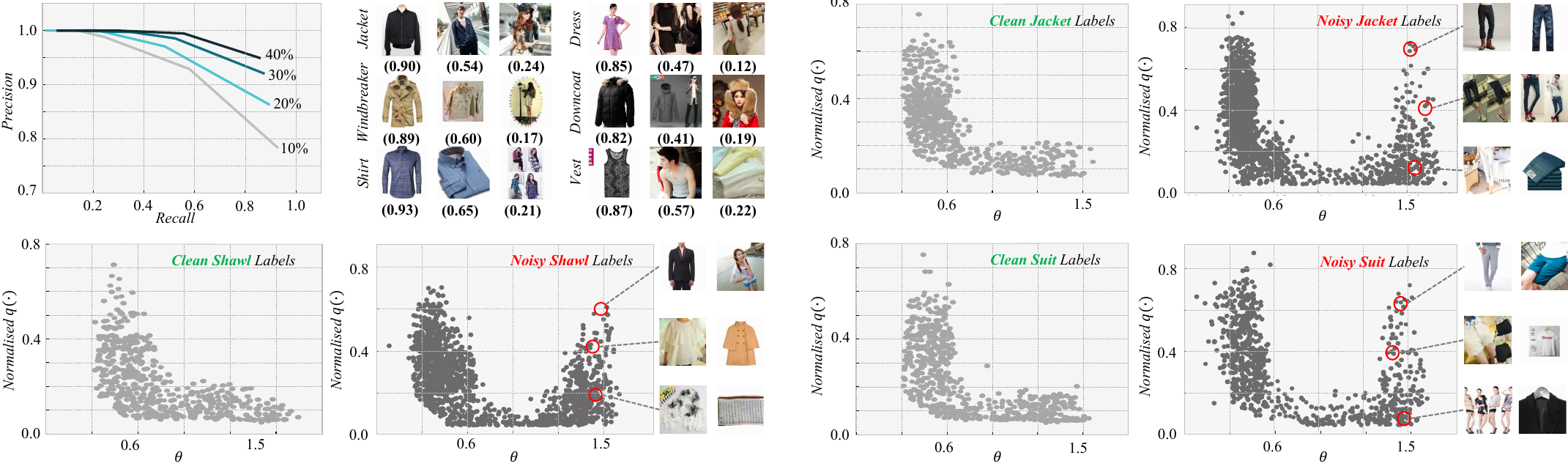}
    \caption{{\textit{Exploiting the feature geometry unique to GACL for automated discovery of noisy labels.} Top left: we show, under the label noise level of $40\%$, GACL can help to identify $80\%$ of erroneous labels with a precision of more than $90\%$. The rest: three noisy label categories, where the distributions of $q(\cdot)$ and $\theta$ follow a ``U'' shape, in contrast to the ``L'' shape formed by clean image labels. This is then exploited for label-cleansing purpose. More details in text.}}
    \label{fig:clothing}
\end{figure*}

\section{Further Analysis}
\label{sec:further}

{We provide further analysis and discussion to show that GACL (i) as a quality discovery method is agnostic to the specific choice of sketch representation and the subsequent network architecture to encode it; (ii) as a data re-weighting method can be generally applied to various vision applications; (iii) GACL has the potential to work for data domain that is not expensively and expertly processed with category label annotations. Below is a detailed description.}

\keypoint{Is GACL constrained to work for vector
+ RNN type of sketch data processing pipeline?} {This paper implements GACL upon a common sketch data learning framework \citep{sarvadevabhatla2016enabling, quickdraw, li2018universal} that feeds a sequence of coordinate vectors into an RNN (BiLSTM) network.
We aim to find out whether GACL can still perform reasonably on quality discovery when sketches are processed as raster format \citep{sketchanet, bhunia2020fgsbir} or orderless point clouds \citep{sketchformer, lin2020sketch}. For raster sketch, we adopt ConvNeXt \citep{liu2022convnet}, a competitive recent CNN variant, as the backbone feature extractor. We conduct global average pooling on the feature maps from its penultimate layer to form $f(x_i)$ in Eq.~\ref{eq:softmax}. Similarly, we adopt Transformer \citep{vaswani2017attention} to encode sketch point clouds. We add an extra learnable ``CLS'' token preceding all sketch points and concatenate its final representation along with the mean representation of all sketch points to form $f(x_i)$. We conduct the same human study procedures as those in Sec.~\ref{sec:humanstudy} and the average human quality agreement percentage of $77.87\%$ and $77.44\%$ with GACL-Transformer and GACL-ConvNeXt (\cf $78.13\%$ with Ours-Cos) confirm GACL's generalisation capability to sketch representation methods.}

\keypoint{Can GACL be a general vision data re-weighting method?} {Essentially, GACL learns to assign a real-valued score to a data instance with the hope of exposing some representative aspect of the training set (\eg different quality levels among sketched bunnies). This concept is largely reminiscent of a commonly adapted machine learning technique of data re-weighting \citep{ren2018learning, dukler2021diva}, which has resulted in many real-world impacts including providing probes into the black-box model dynamics \citep{koh2017understanding} and tackling the noisy label or class imbalance problems \citep{shu2019meta, cui2019class}. We here explore whether GACL, can set its feet outside the sketch quality problem domain, and cater to a broader range of end vision tasks as a more general data re-weighting method.}

 \doublecheck{We leverage GACL for a common problem of label cleansing on Clothing1M\footnote{\doublecheck{In Appendix, we showcase more applications of GACL as a general data reweighting method, including filtering out ambiguous and destructive benchmark data and withstanding an ethical check when using face recognition as an example}}.} For that, we train GACL on the Clothing1M dataset \citep{clothing1m}, which contains more than a million images crawled from several online shopping websites with their labels converted from the accompanying texts and tags. These labels from 14 classes (T-shirt, Knitwear, Sweater, Hoodie, Windbreaker, $\cdots$) are thus inherently noisy. The dataset manually refined a small proportion of validation images by rectifying the potentially erroneous labels, of which we choose 1,000 random samples for each class. We then inject label noise to each data instance (changing the label to one of the other 13 classes) with chances ranging from $0\%$ to $40\%$ and want to find out whether GACL can help to identify those tampered labels. The key insight here is that with the right data-label synergy, $q(\cdot)$ and $\theta$ follow \doublecheck{an} ``L-shaped'' distribution (Fig.~\ref{fig:relation}) (\doublecheck{the higher quality sample should have a smaller geometric distance to its class centre}). In contrast, data with a significant portion of noisy labels will stretch that distribution into ``U-shape''  as shown in Fig.~\ref{fig:clothing}. It can be seen that error-prone labels mainly reside at a self-organising, crowded region characterised by \textit{``large q and large $\theta$''}, \ie \doublecheck{region where high quality samples exhibit an unusually large geometric distance from their associated class centres}. We regard samples with \doublecheck{$\theta$} larger than a threshold value $1.5$ and quality scores larger than $0.4$ as the ones erroneously labelled. We plot the precision-recall curve in Fig.~\ref{fig:clothing}. It can be seen that precision remains around $90\%$ even when $80\%$ of the samples with noisy labels get recalled. Given the average occupation rate $38.36\%$ of noisy labels in Clothing1M roughly estimated by the original paper \citep{clothing1m}, such a result is stimulating as it suggests an ideal scenario of converting Clothing1M to a much cleaner dataset with a label corruption rate less than $10\%$, all achieved without expensive human intervention.

\begin{figure*}
    \centering
    \includegraphics[width=\linewidth]{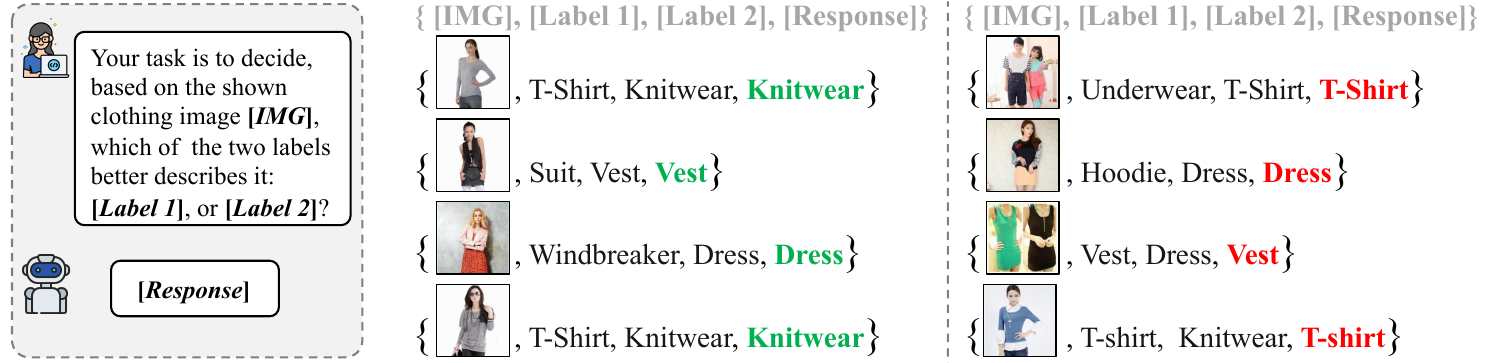}
    \caption{\textit{LLaVA as free and scalable annotator.} \doublecheck{Given an image query [IMG], we ask LLaVA to make a binary choice between the label originally attached with the image ([Label 1]) and the label deemed more proper by GACL ([Label 2]). We highlight the [response] of LLaVA, using ``green'' to speak for cases where LLaVA's response is endorsed by human inspection later and using ``red'' otherwise.}}
    \label{fig:clothing_result}
\end{figure*}

\doublecheck{We also conduct label cleansing experiment directly on the entire Clothing1M benchmark with the setting ``$q_i > 0.4$ \& $\theta_i > 1.5$'' and ``$q_i < 0.1$ (image quality is too bad to be properly labelled)''. Overall, 35.14\% samples (3,514 out of 10,000) are deemed as wrongly labelled, which is close to the number 38.36\% estimated by the original paper. To really look into the correctness of this number is however extremely challenging, given we don't have the budget to annotate them all. We have since done two things striving to shed light on the efficacy of label cleansing under our proposal: i) we first use a powerful multi-modal foundation model such as LLaVA as a more affordable and scalable data annotator \citep{liu2023visual}. Despite its remarkable generalisation ability, LLaVA has shown several fatal weaknesses towards responsible and practical adoption, with confabulation or hallucination being the most famous \citep{chiang2023can}. We find that directly asking LLaVA to annotate the labels for images is highly unreliable due to hallucinations \citep{li2023evaluating}, especially in cases of true noisy labels where images are often less typical (a woman in side view wearing a shawl). Thus instead of prompting LLaVA to return us one of fourteen labels formulated in Clothing1M, we portray the role of LLaVA as a judge -- so that given an image-label pair deemed as un-matching by our proposed method, we ask LLaVA to review this decision and give its own say. In total, LLaVA endorses 70.76\% of the cases where our proposed method believes there are wrong/noisy labels taking place. ii) we conduct a mid-scale human study to examine the reliability of LLaVA. We recruit 5 participants and task them with the same question as set for LLaVA. In a total of 200 trials, humans at 78.5\% of the time find themselves in alignment with LLaVA choice. For cases where LLaVA is making mistakes, they also feel understandable for many of them (53.4\%) where images could present more than one clothing item or even more than one person -- thus labelling them becomes an ill-posed problem itself. Some typical annotations (including right and wrong) by LLaVA are illustrated in Fig.\ref{fig:clothing_result}.}

\begin{figure}[t]
    \centering
    \includegraphics[width=\linewidth]{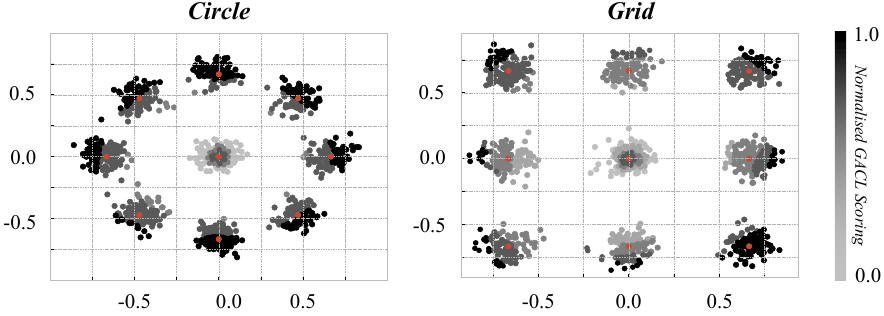}
    \caption{{\textit{GACL works for pseudo labels.} We apply GACL on two synthetic toy datasets, where each 2D data point comes with zero annotation, even without the one-hot category label we assumed through the paper. We then generate offline pseudo labels for GACL in a way similar to unsupervised clustering and show some promising results under this extreme context.}}
    \label{fig:no_label}
\end{figure}

\keypoint{Can GACL adapt to data regime without category label annotations?} {So far, GACL relies on the availability of annotations of category labels in order to perform quality scoring. This is not necessarily a downside as we have already confirmed that GACL is able to work on image datasets (Clothing1M) with labels ``freely'' crawled from web-rich texts. That said, we still pose the question of whether GACL can cope with the extreme scenario of zero annotation availability and conduct a pilot study using toy examples. We train GACL with a simple backbone of a four-layer perceptron on a 2D mixture of 9 Gaussians ($\sigma$=0.05) arranged in a circle or 9 Gaussians arranged in an equally-spaced grid. Pseudo labels required by GACL are generated and updated every a few training epochs in a similar fashion to DeepCluster \citep{caron2018deep}, where each 2D point is assigned with the label of the class centres closest to its feature representation. We demonstrate the scatter plot of GACL scorings in Fig.~\ref{fig:no_label} and can observe that GACL works nicely by assigning higher scores to data points that keep safer (farther) distance from the decision boundary, \ie points that are more representative of their Gaussian centre. It is interesting to see how the data points sourced from the centred Gaussian distribution are generally scored unfavourably by GACL. One way to interpret is that they can be considered less individualistic/distinctive when the rest of the data space is revolving around and thus sharing a connection with them.
} 

\section{Conclusion}

We have presented a method for quantifying human free-hand sketch quality. Without relying on supervision from human quality opinion annotations for learning, our proposed solution GACL is able to stand up to the test of a large scale human study by showing human-agreeable results on sketch quality discrimination. We also demonstrate three practical sketch use cases benefited from successful sketch quality modelling, and the broader scope of vision applications GACL potentially enables. Last but not least, we hope our work can be generally helpful to sketch practitioners, who seek to differentiate good/bad bunny drawings in their model deployments but are held back by the lack of a proper sketch quality metric today.

\section*{Declarations} 

\textbf{Conflict of interest} The authors have no competing interests to declare that are relevant to the content of this article.

\backmatter

%\bibliographystyle{sn-mathphys.bst}

\bibliography{egbib}% common bib file
%\input sn-article.bbl

%\input sn-sample-bib.tex

\end{document}